\documentclass[10pt,twocolumn,letterpaper]{article}

\usepackage[pagenumbers]{cvpr} % To force page numbers, e.g. for an arXiv version

\usepackage{graphicx}
\usepackage{amsmath}
\usepackage{amssymb}
\usepackage{booktabs}

\usepackage{array,multirow}

\usepackage[dvipsnames]{xcolor}

\usepackage[accsupp]{axessibility}
\usepackage[pagebackref,breaklinks,colorlinks]{hyperref}

\usepackage[capitalize]{cleveref}
\crefname{section}{Sec.}{Secs.}
\Crefname{section}{Section}{Sections}
\Crefname{table}{Table}{Tables}
\crefname{table}{Tab.}{Tabs.}

\newcommand\blue[1]{\textcolor{blue}{#1}}
\newcommand\red[1]{\textcolor{red}{#1}}

\newcommand\green[1]{\textcolor{OliveGreen}{\textbf{#1}}}
\newcommand\maroon[1]{\textcolor{Maroon}{\textbf{#1}}}

\def\cvprPaperID{2882} % *** Enter the CVPR Paper ID here
\def\confName{CVPR}
\def\confYear{2023}

\begin{document}

\title{Learned Two-Plane Perspective Prior based Image Resampling for \\Efficient Object Detection}

\author{Anurag Ghosh \qquad N. Dinesh Reddy
\qquad Christoph Mertz
\qquad Srinivasa G. Narasimhan \\
Carnegie Mellon University\\
{\tt\small {\{anuraggh, dnarapur, cmertz, srinivas\}}@cs.cmu.edu}\\
}
\maketitle

%%%%%%%%% ABSTRACT
\begin{abstract}
\noindent
Real-time efficient perception is critical for autonomous navigation and city scale sensing. Orthogonal to architectural improvements, streaming perception approaches have exploited adaptive sampling improving real-time detection performance. In this work, we propose a learnable geometry-guided prior that incorporates rough geometry of the 3D scene (a ground plane and a plane above) to resample images for efficient object detection. This significantly improves small and far-away object detection performance while also being more efficient both in terms of latency and memory. For autonomous navigation, using the same detector and scale, our approach improves detection rate by \green{+4.1 $AP_{S}$} or \textbf{+$39\%$} and in real-time performance by \green{+5.3 $sAP_{S}$} or \textbf{+$63\%$} for small objects over state-of-the-art (SOTA). For fixed traffic cameras, our approach detects small objects at image scales other methods cannot. At the same scale, our approach improves detection of small objects by \textbf{$195\%$} (\green{+12.5 $AP_{S}$}) over naive-downsampling and \textbf{$63\%$} (\green{+4.2 $AP_S$}) over SOTA.
\end{abstract}

%%%%%%%%% BODY TEXT
\section{Introduction}
\label{sec:intro}

\noindent
Visual perception is important for autonomous driving and decision-making for smarter and sustainable cities. Real-time efficient perception is critical to accelerate these advances. For instance, a single traffic camera captures half a million frames every day or a commuter bus acting as a city sensor captures one million frames every day to monitor road conditions~\cite{christensen2019towards} or to inform public services~\cite{hull2006cartel}. There are thousands of traffic cameras~\cite{vox2015trafficcams} and nearly a million commuter buses~\cite{manybuses} in the United States. It is infeasible to transmit and process visual data on the cloud, leading to the rise of edge architectures~\cite{satyanarayanan2017emergence}. However, edge devices are severely resource constrained and real-time inference requires down-sampling images to fit both latency and memory constraints severely impacting accuracy. 

\begin{figure}
    \centering
    \includegraphics[width=1\linewidth]{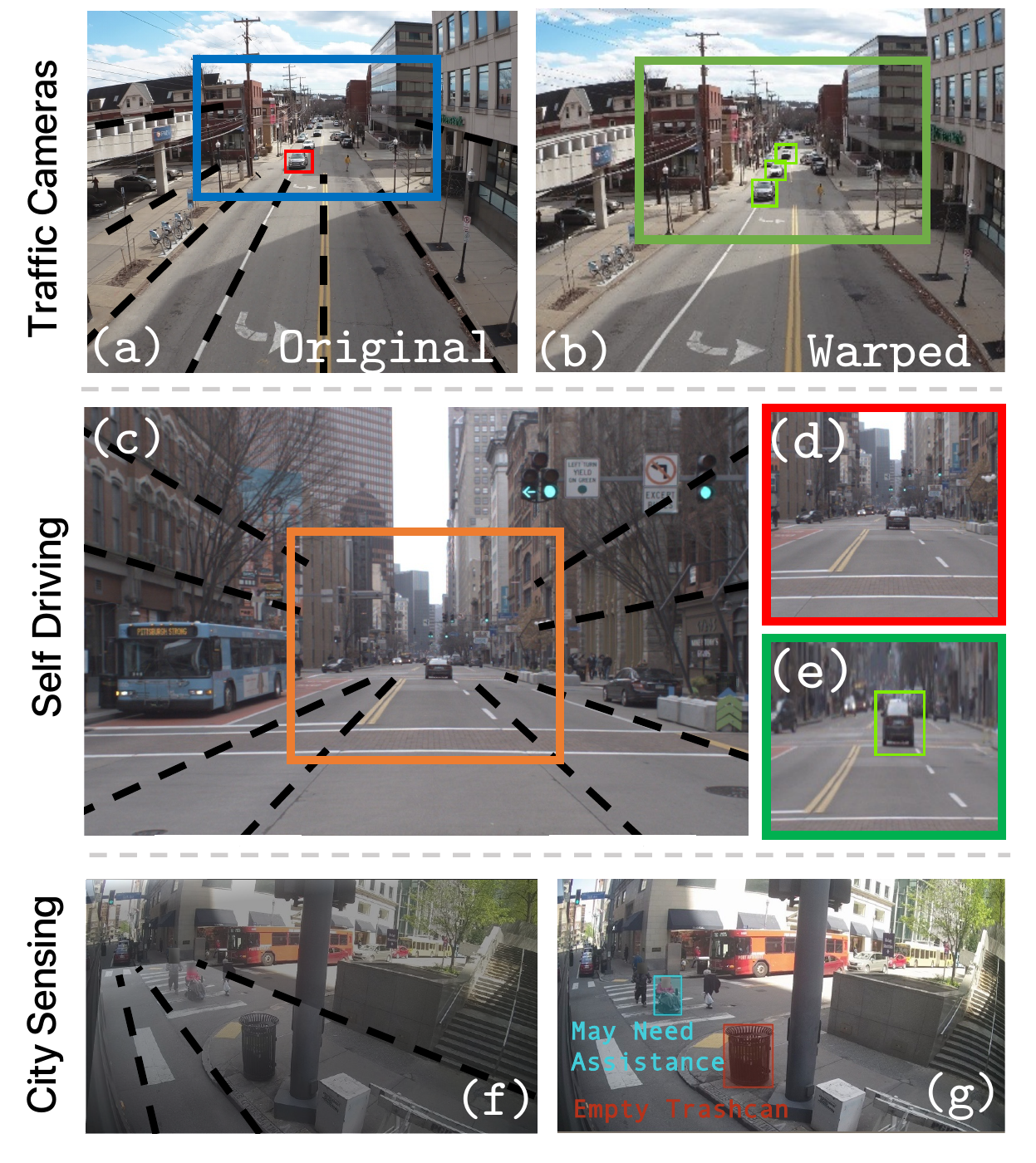}
    \caption{Geometric cues (black dashed lines) are implicitly present in scenes. Our Perspective based prior exploits this geometry. Our method (a) takes an image and (b) warps them, and performs detection on warped images. Small objects which are (d) not detected when naively downsampled but (e) are detected when enlarged with our geometric prior. Our method (f) uses a geometric model to construct a saliency prior to focus on relevant areas and (g) enables sensing on resource-constrained edge devices.}
    \label{fig:teaser}
    \vspace{-0.2in}
\end{figure}

\noindent
On the other hand, humans take visual shortcuts~\cite{hayes2019scene} to recognize objects efficiently and employ high-level semantics~\cite{hayes2019scene, torralba2006contextual} rooted in scene geometry to focus on relevant parts. Consider the scene in Figure~\ref{fig:teaser} (c), humans can recognize the distant car despite its small appearance (Figure~\ref{fig:teaser} (d)). We are able to contextualize the car in the 3D scene, namely (1) it's on the road and (2) is of the right size we'd expect at that distance. Inspired by these observations, can we incorporate semantic priors about scene geometry in our neural networks to improve detection? 

\noindent
In this work, we develop an approach that enables object detectors to ``zoom'' into relevant image regions (Figure~\ref{fig:teaser} (d) and (e)) guided by the geometry of the scene.  Our approach considers that most objects of interests are present within two planar regions, either on the ground plane or within another plane above the ground, and their size in the image follow a geometric relationship. Instead of uniformly downsampling, we sample the image to enlarge far away regions more and detect those smaller objects.

\noindent
While methods like quantization~\cite{gupta2015deep}, pruning~\cite{han2015learning}, distillation~\cite{chen2017learning} and runtime-optimization~\cite{ghosh2021adaptive} improve model efficiency (and are complementary), approaches exploiting spatial and temporal sampling are key for enabling efficient real-time perception~\cite{huang2017speed, li2020towards}. Neural warping mechanisms~\cite{jaderberg2015spatial, recasens2018learning} have been employed for image classification and regression, and recently, detection for self-driving~\cite{thavamani2021fovea}. Prior work~\cite{thavamani2021fovea} observes that end-to-end trained saliency networks fail for object detection. They instead turn to heuristics such as dataset-wide priors and object locations from previous frames, which are suboptimal. We show that formulation of learnable geometric priors is critical for learning end-to-end trained saliency networks for detection.

\noindent
We validate our approach in a variety of scenarios to showcase the generalizability of geometric priors for detection in self-driving on Argoverse-HD~\cite{li2020towards} and BDD100K~\cite{yu2020bdd100k} datasets, and for traffic-cameras on  WALT~\cite{reddy2022walt} dataset. 
\begin{itemize} \itemsep -0.25em
    \item On Argoverse-HD, our learned geometric prior improves performance over naive downsampling by \green{+6.6 $AP$} and \green{+2.7 $AP$} over SOTA using the same detection architecture. Gains from our approach are achieved by detecting small far-away objects, improving by \green{9.6 $AP_S$} (or $195\%$) over naive down-sampling and \green{4.2 $AP_S$} (or $63\%$) over SOTA.
    \item On WALT, our method detects small objects at image scales where other methods perform poorly. Further, it significantly improves detection rates by \green{10.7 $AP_{S}$} over naive down-sampling and \green{3 $AP_{S}$} over SOTA.
    \item Our approach improves object tracking  (+$4.8\%$ MOTA) compared to baseline. It also improves tracking quality, showing increase of \green{+7.6\% $MT\%$} and reduction of \green{-6.7\% $ML\%$}. 
    \item  Our approach can be deployed in resource constrained edge devices like Jetson AGX to detect $42\%$ more rare instances while being \green{2.2X} faster to enable real-time sensing from buses.
\end{itemize}

\section{Related Work}
\label{sec:related}
\noindent
We contextualize our work with respect to prior works modelling geometry and also among works that aim to make object detection more accurate and efficient.

\noindent
\textbf{Vision Meets Geometry:} Geometry has played a crucial role in multiple vision tasks like detection~\cite{hoiem2008putting, sudowe2011efficient, wang2019pseudo, chen20153d}, segmentation~\cite{sturgess2009combining, li2017foveanet}, recognition~\cite{Geiger2014PAMI, su2015multi} and reconstruction~\cite{schoenberger2016sfm, 3DRCNN_CVPR18, Narapureddy-2018-105893}. Perspective Geometric constraints have been used to remove distortion~\cite{zhao2019learning}, improve depth prediction and semantic segmentation~\cite{ladicky2014pulling} and feature matching~\cite{toft2020single}. However, in most previous works~\cite{wang2019pseudo, schoenberger2016sfm, 3DRCNN_CVPR18} exploiting these geometric constraints have mainly been concentrated around improving 3D understanding. This can be attributed to a direct correlation between the constraints and the accuracy of reconstruction. Another advantage is the availability of large RGB-D and 3D datasets~\cite{chang2015shapenet, Geiger2012CVPR, reizenstein2021common} to learn and exploit 3D constraints. Such constraints have been under-explored for learning based vision tasks like detection and segmentation. 
A new line of work interpreting classical geometric constraints and algorithms as neural layers~\cite{rockwell20228, chen2022epro} have shown considerable promise in merging geometry with deep learning.  

\noindent
\textbf{Learning Based Detection:} Object detection has mostly been addressed as an learning problem. Even classical-vision based approaches~\cite{dalal2005histograms, viola2001rapid} extract image features and learn to classify them into detection scores. With deep learning, learnable architectures have been proposed following this paradigm~\cite{ren2015faster, redmon2016you, carion2020end, liu2021swin}, occasionally incorporating classical-vision ideas such as feature pyramids for improving scale invariance~\cite{lin2017feature}. While learning has shown large improvements in accuracy over the years they still perform poorly while detecting small objects due to lack of geometric scene understanding. To alleviate this problem, we guide the input image with geometry constraints, and our approach complements these architectural improvements.  

\noindent
\textbf{Efficient Detection with Priors:} Employing priors with learning paradigms achieves improvements with little additional human labelling effort. Object detection has traditionally been tackled as a learning problem and geometric constraints were sparsely used for such tasks, constraints like ground plane~\cite{hoiem2008putting, sudowe2011efficient} were used.  

\noindent
Temporality~\cite{ehteshami2022salisa, thavamani2021fovea, yang2022streamyolo} has been exploited for improving detection efficiently. Some of these methods~\cite{ehteshami2022salisa, thavamani2021fovea} deform the input image using approach that exploit temporality to obtain saliency. This approach handles cases where object size decreases with time (object moving away from the camera in scene), but cannot handle new incoming objects. None of these methods explicitly utilize geometry to guide detection, which handles both these cases. Our two-plane prior deforms the image while taking perspective into account without biasing towards previous detections.

\noindent
Another complementary line of works automatically learn metaparameters (like image scale)~\cite{ghosh2021adaptive, sela2022context, chin2019adascale} from image features. However, as they do not employ adaptive sampling accounting for image-specific considerations, performance improvements are limited. Methods not optimized for online perception like AdaScale~\cite{chin2019adascale} for video object detection do not perform well in real-time situations.

\section{Approach}
\label{sec:approach}
\noindent
We describe how a geometric model rooted in the interpretation of a 3D scene can be derived from the image. We then describe how to employ this rough 3D model to construct saliency for warping images and improving detection. 

\subsection{Overview}
\label{sec:overview}

\noindent
Object sizes in the image are determined by the 3D geometry of the world. Let us devise a geometric inductive prior considering a camera mounted on a vehicle. Without loss of generality, assume the vehicle is moving in direction of the dominant vanishing point.

\noindent
We are interested in objects that are present in a planar region (See Figure~\ref{fig:parameterization}) of width $P_{1}P_{2}$ corresponding to the camera view, of length $P_{1}P_{3}$ defined in the direction of the vanishing point. This is the planar region on the ground on which most of the objects of interest are placed (vehicles, pedestrians, etc) and another planar region $Q_{1} ... Q_{4}$ parallel to this ground plane above horizon line, such that all the objects are within this region (e.g., traffic lights).

\noindent
From this simple geometry model, we shall incorporate relationships derived from perspective geometry about objects, i.e., the scale of objects on ground plane is inversely proportional to their depth w.r.t camera~\cite{hoiem2008putting}. 

\subsection{3D Plane parameterization from 2D images}
\label{sec:3dreasoning}

\begin{figure}
    \centering
    \includegraphics[width=0.7\linewidth]{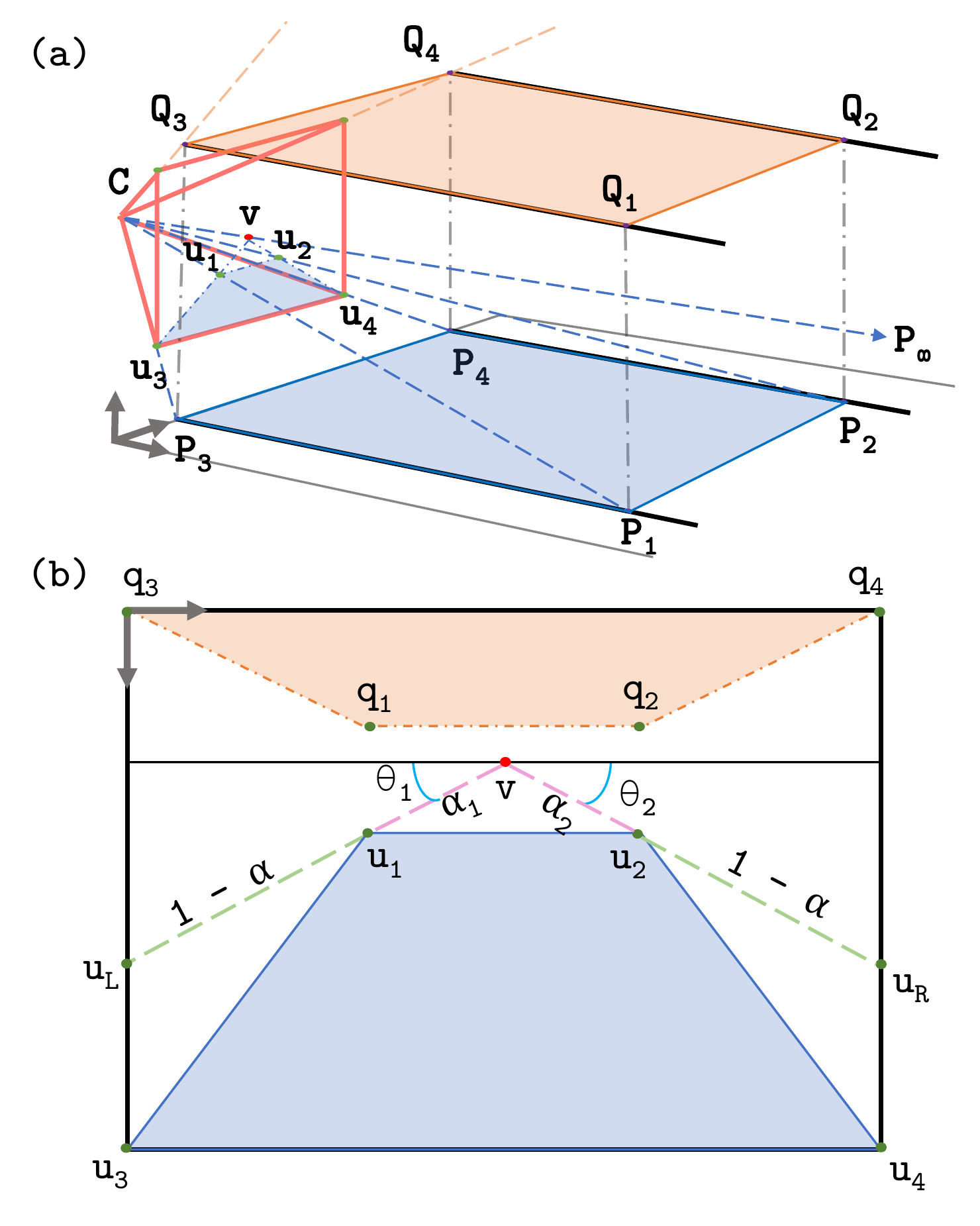}
    \caption{\textbf{Geometry Of The Two Plane Perspective Prior:}  (a) describes the single view geometry of the proposed two plane prior. Region on the ground plane defined by $P_{1}, ... P_{4}$, and rays emanating from camera $C$ to $P_{i}$ intersect at $u_{1} ... u_{4}$ on the image plane. The vanishing point $v$ maps to $P_{\infty}$. This planar region accounts for small objects on the ground plane. To account for objects that are tall or do not lie on the ground plane, we consider another plane $Q_{1} .. Q_{4}$ above the horizon line. These two planes encapsulate all the relevant objects in the scene. (b) depicts the re-parameterization of the two planes in the 2D image. Instead of representing the planar points $u_{1} ... u_{4}$ as pixel coordinates, we instead parameterize them in terms of the vanishing point $v$, $\theta$'s and $\alpha$ to ease learning.}
    \label{fig:parameterization}
\end{figure}

\noindent
We parameterize the planes of our inductive geometric prior. We represent 2D pixel projections $u_{1} ... u_{4}$ of 3D points $P_{1} ... P_{4}$. Assume that the dominant vanishing point in the image is $v = (v_{x}, v_{y})$ and let the image size be $(w, h)$. Consider $u_{1}$ (Figure~\ref{fig:parameterization} (b)). We can define a point on the edge of the image plane,

\vspace{-0.1in}
\begin{equation}
    \label{eq:pleft}
    u_{L} = (0, v_{y} + v_{x}\tan{\theta_{1}})   
\end{equation}

\noindent
$u_{1}$ can expressed as a linear combination of $v$ and $u_{L}$,

\vspace{-0.1in}
\begin{equation}
    \label{eq:p1}
    u_{1} = \alpha_{1} u_{L} + (1 - \alpha_{1}) v
\end{equation}

\noindent
Similarly, for $u_{2}$, we can define $u_{R}$ in terms of $v$ and $\theta_{2}$ and $\alpha_{2}$ while $u_{3}$ and $u_{4}$ are defined like Equation~\ref{eq:pleft} to represent any arbitrary plane in this viewing direction. However, for simplicity, for ground plane we fix them as $(0, h)$ and $(w, h)$ respectively. Consider the planar region $Q_{1} ... Q_{4}$ at height $H$ above the horizon line.  We can similarly define $\theta_{3}$ and $\theta_{4}$ to represent the angles from the horizon in the opposite direction and define $q_{1}$ and $q_{2}$. Again, we set $q_{3}$ as $(0, 0)$ and $q_{4}$ as $(w, 0)$. We now have $4$ points to calculate homographies $H_{plane}$ for both  planes. 

\begin{figure*}[t]
    \centering
    \includegraphics[width=1\linewidth]{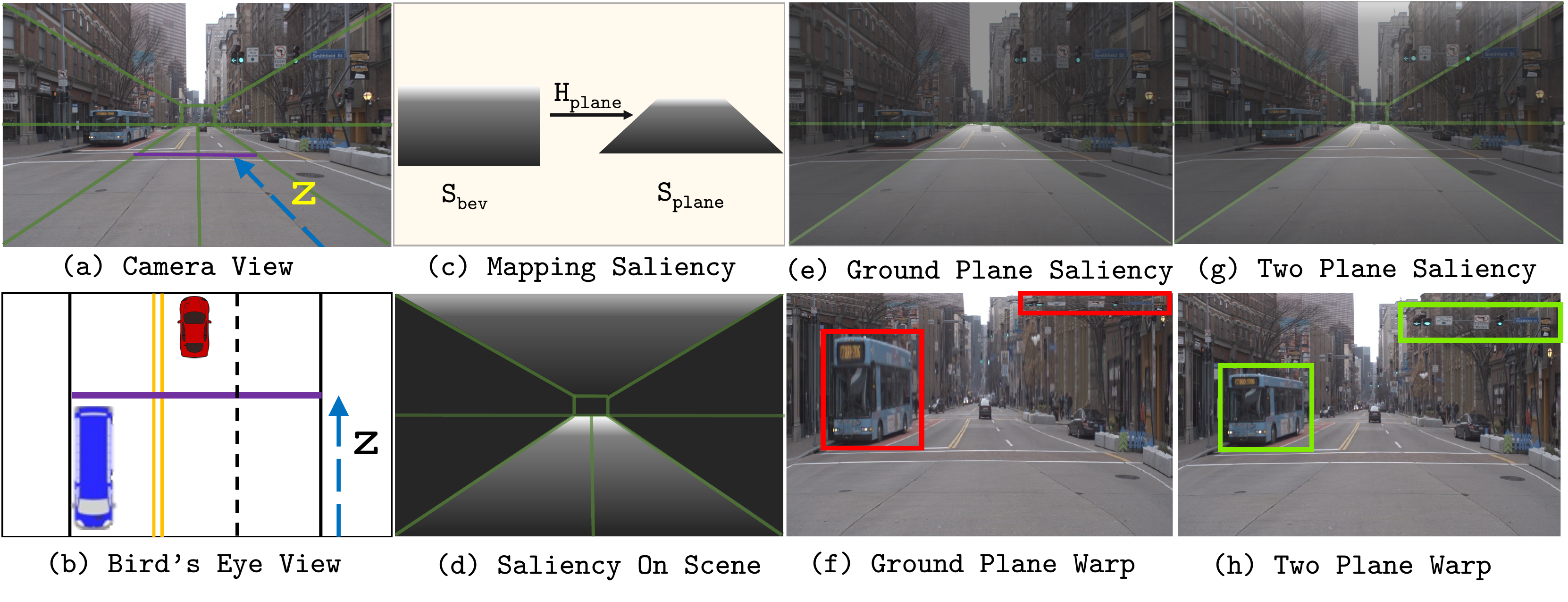}
    \caption{\textbf{Two-Plane Perspective Prior based Image Resampling:} Consider the scene of car, bus and traffic light from (a) camera view and (b) (simplified) bird's eye view. (c) Saliency function that captures the inverse relationship between object size (in camera view) and depth (bird's eye view is looking at $XZ$ plane from above) can be transferred to the camera view (d), by mapping row $z$ using $H$ (marked by blue arrows). (e) and (f) shows that ground plane severely distorts nearby tall objects while squishing traffic light. (g) and (h) shows that additional plane reduces distortion for both tall objects and objects not on ground plane.}
    \label{fig:saliency-explanation}
\end{figure*}

\noindent
For now, assume $v$ is known. However, we still do not know the values for $\theta$'s and $\alpha$, and we shall learn these parameters end-to-end from task loss. These parameters are learned and fixed for a given scenario in our learning paradigm. Our re-parameterization aims to ease learning of these parameters as we clamp the values of $\alpha$'s to $[0, 1]$ and $\theta$'s to $[-\frac{\pi}{2}, \frac{\pi}{2} ]$. It should be noted that all the operations are differentiable.

\subsection{From Planes to Saliency}
\label{sec:avoiding-cropping}

\noindent
We leverage geometry to focus on relevant image regions through saliency guided warping~\cite{jaderberg2015spatial, recasens2018learning}, and create a saliency map from a parameterized homography using $u_{1} .. u_{4}$ defined earlier. Looking at ground plane from two viewpoints (Figure~\ref{fig:saliency-explanation} (a) and (b)), object size decreases by their distance from the camera~\cite{hoiem2008putting}. We shall establish a relationship to counter this effect and ``sample'' far-away objects on the ground plane more than nearby objects.

\noindent
The saliency guided warping proposed by~\cite{recasens2018learning} operates using an inverse transformation $\mathcal{T}_{S}^{-1}$ parameterized by a saliency map $S$ as follows,

\vspace{-0.1in}
\begin{equation}
    I'(x, y) = W_{\mathcal{T}}(I) = I(\mathcal{T}_{S}^{-1}(x, y))
\end{equation}

\noindent
where the warp $W_{\mathcal{T}}$ implies iterating over output pixel coordinates, using \textbf{$\mathcal{T}_{S}^{-1}$} to find corresponding input coordinates (non-integral), and bilinearly interpolating output color from neighbouring input pixel grid points. For each input pixel $(x, y)$, pixel coordinates with higher $S(x, y)$ values (i.e. salient regions) would be sampled more. 

\noindent
We construct a saliency $S$ respecting the geometric properties that we desire. Let $H_{plane}$ be the homography between the camera view (using coordinates $u_{1} ... u_{4}$) and a bird's eye view of the ground plane assuming plane size to be the original image size $(w, h)$. In bird's eye view, we propose saliency function for a row of pixels $z$ (assuming bottom-left of this rectangle as (0, 0)) as,

\vspace{-0.1in}
\begin{equation}
    \label{eq:saliency}
    S_{bev}(z) = e^{\nu(\frac{z}{h} - 1)}
\end{equation}

\noindent
with a learnable parameter $\nu$ ($> 1$). $\nu$ defines the extent of sampling with respect to depth. 

\noindent
To map this saliency to camera view, we warp $S_{bev}$ via perspective transform $W_{p}$ and $H_{plane}$ (Figure~\ref{fig:saliency-explanation} (c)),

\vspace{-0.1in}
\begin{equation}
    \label{eq:saliency-warps}
    S_{plane} = W_{p}(H^{-1}_{plane}, S_{bev})
\end{equation} 

\noindent
We have defined saliency $S_{plane}$ given $H_{plane}$ in a differentiable manner. Our saliency ensures that objects on the ground plane separated by depth $Z$ are sampled by the factor $e^{\nu \frac{Z}{h}}$ in the image.

\subsection{Two-Plane Perspective Prior}
\label{sec:complete-prior}

\noindent
Ground Plane saliency focuses on objects that are geometrically constrained to be on this plane and reasonably models objects far away on the plane. However, nearby and tall objects, and small objects far above the ground plane are not modelled well. In Fig~\ref{fig:saliency-explanation} (f), \textit{nearby objects above ground plane} (traffic lights), they are highly distorted. Critically, these \textit{same objects when further away} are rendered small in size and appear close to ground (and thus modelled well). Objects we should \textbf{focus} more on are thus the former compared to the latter. Thus, another plane is needed, and direction of the saliency function is reversed to $\hat{S}_{bev}(z) = e^{\hat{\nu}(((h - z)/h) - 1)}$ to account for these objects that would otherwise be severely distorted.

\noindent
To represent the Two-Plane Prior, we represented the planar regions as saliencies. The overall saliency is,

\vspace{-0.1in}
\begin{equation}
    S = S_{ground\_plane} + \lambda S_{top\_plane}
\end{equation}

\noindent
where $\lambda$ is a learned parameter.

\subsection{Additional Considerations}

\noindent
Warping via a piecewise saliency function imposes additional considerations. The choice of deformation method is critical, saliency sampler~\cite{recasens2018learning} implicitly avoids drastic transformations common in other appraoches. For e.g., Thin-plate spline performs worse~\cite{recasens2018learning}, produces extreme transformations and requires regularization~\cite{ehteshami2022salisa}.

\noindent
Fovea~\cite{thavamani2021fovea} observes that restricting the space of allowable warps such that axis alignment is preserved improves accuracy, we adopt the separable formulation of $\mathcal{T}^{-1}$,    

\vspace{-0.1in}
\begin{equation}
    \mathcal{T}_{x}^{-1}(x) = \frac{\int_{x'} S_{x}(x') k(x', x) x'}{\int_{x'} S_{x}(x') k(x, x')}
\end{equation}

\begin{equation}
    \mathcal{T}_{y}^{-1}(y) = \frac{\int_{y'} S_{y}(y') k(y', y) y'}{\int_{y'} S_{y} (y') k(y, y')}
\end{equation}

\noindent
where $k$ is a Gaussian kernel. To convert a saliency map $S$ to $S_{x}$ and $S_{y}$ we marginalize it along the two axes. Thus entire rows or columns are ``stretched'' or ``compressed''.

\noindent
Two-plane prior is learnt end to end as a learnable image warp. For object detection, labels need to be warped too, and~\cite{recasens2018learning}'s warp is invertible. Like~\cite{thavamani2021fovea}, We employ the loss $\mathcal{L}(\mathcal{T}^{-1}(f_{\phi}(W_{\mathcal{T}}(I)), L)$  where $(I, L)$ is the image-label pair and omit the use of delta encoding for training RPN~\cite{ren2015faster} (which requires the existence of a closed form $\mathcal{T}$), instead adopting GIoU loss~\cite{rezatofighi2019generalized}. This ensures $W_{\mathcal{T}}$ is learnable, as $\mathcal{T}^{-1}$ is differentiable.

\noindent
We \textit{did not} assume that the vanishing point is within the field of view of our image, and our approach places no restrictions on the vanishing point. Thus far, we explained our formulation while considering a single dominant vanishing point, however, multiple vanishing points can be also considered. Please see supplementary for more details.

\subsection{Obtaining the Vanishing Point}

\noindent
We now describe how we obtain the vanishing point. Many methods exist with trade-offs in accuracy, latency and memory which which inform our design to perform warping efficiently with minimal overheads.

\noindent
\textbf{Fixed Cameras:} In settings like traffic cameras, the camera is fixed. Thus, the vanishing point is fixed, and we can cache the corresponding saliency $S$, as all the parameters, once learnt, are fixed. We can define the vanishing point for a camera manually by annotating two parallel lines or any accurate automated approach. Saliency caching renders our approach extremely efficient.

\noindent
\textbf{Autonomous Navigation:} 
Multiple assumptions simplify the problem. We assume that there is one dominant vanishing point and a navigating car is often moving in the viewing direction. Thus, we assume that this vanishing point lies inside the image, and directly regress $v$ from image features using a modified coordinate regression module akin to YOLO~\cite{redmon2016you, liu2020d}. This approach appears to be memory and latency efficient. Other approaches, say, using parallel lane lines~\cite{abualsaud2021laneaf} or inertial measurements~\cite{camposeco2015using} might also be very efficient. An even simpler assumption is to employ the average vanishing point, as vanishing points are highly local, we observe this is a good approximation.

\noindent
\textbf{Temporal Redundancies:} In videos, we exploit temporal redundancies, the vanishing point is computed every $n_{v}$ frames and saliency is cached to amortize latency cost.

\noindent
\textbf{General Case:} This is the most difficult case, and many approaches have been explored in literature to find all vanishing points. Classical approaches~\cite{tardif2009non} while fast are not robust, while deep-learned approaches~\cite{zhou2019neurvps, lin2022deep, liu2021vapid} are accurate yet expensive (either in latency or memory).

\subsection{Learning Geometric Prior from Pseudo-Labels}
\label{sec:pseudo-labels}

\noindent
Prior work~\cite{thavamani2021fovea} have shown improvements in performance on pre-trained models via heuristics, which didn't require any training. However, their method \textit{still employs domain-specific labels} (say, from Argoverse-HD) to generate the prior. Our method can't be used directly with pre-trained models as it learns geometrically inspired parameters end-to-end. However, domain-specific images without labels can be exploited to learn the parameters.

\noindent
We propose a simple alternative to learn the proposed prior using pre-trained model without requiring additional domain-specific labels. We generate pseudo-labels from our pre-trained model inferred at 1x scale. We fine-tune and learn our warp function (to 0.5x scale) end-to-end using these ``free'' labels. Our geometric prior shows improvements without access to ground truth labels.

\section{Dataset And Evaluation Details}
\label{sec:experiments}
\subsection{Datasets}
\label{sec:datasets}

\noindent
\textbf{Argoverse-HD~\cite{li2020towards}:} We employ Argoverse-HD dataset for evaluation in the autonomous navigation scenario. This dataset consists of 30fps video sequences from a car collected at $1920\times1200$ resolution, and dense box annotations are provided for common objects of interest such as vehicles, pedestrians, signboards and traffic lights. 

\noindent
\textbf{WALT~\cite{reddy2022walt}:} We employ images from $8$ 4K cameras that overlook public urban settings to analyze the flow of traffic vehicles. The data is captured for 3-second bursts every few minutes and only images with notable changes are stored. We annotated a set of 4738 images with vehicles collected over the time period of a year covering a variety of day/night/dawn settings, seasonal changes and camera viewpoints. We show our results on two splits (approximately 80\% training and 20\% testing). The first, \textbf{All-Viewpoints}, images from all the cameras are equally represented in the train and test sets. Alternatively, we split by camera, \textbf{Split-by-Camera}, images from 6 cameras are part of the training set and 2 cameras are held out for testing. 

\begin{figure}[t!]
    \centering
    \includegraphics[width=1\linewidth]{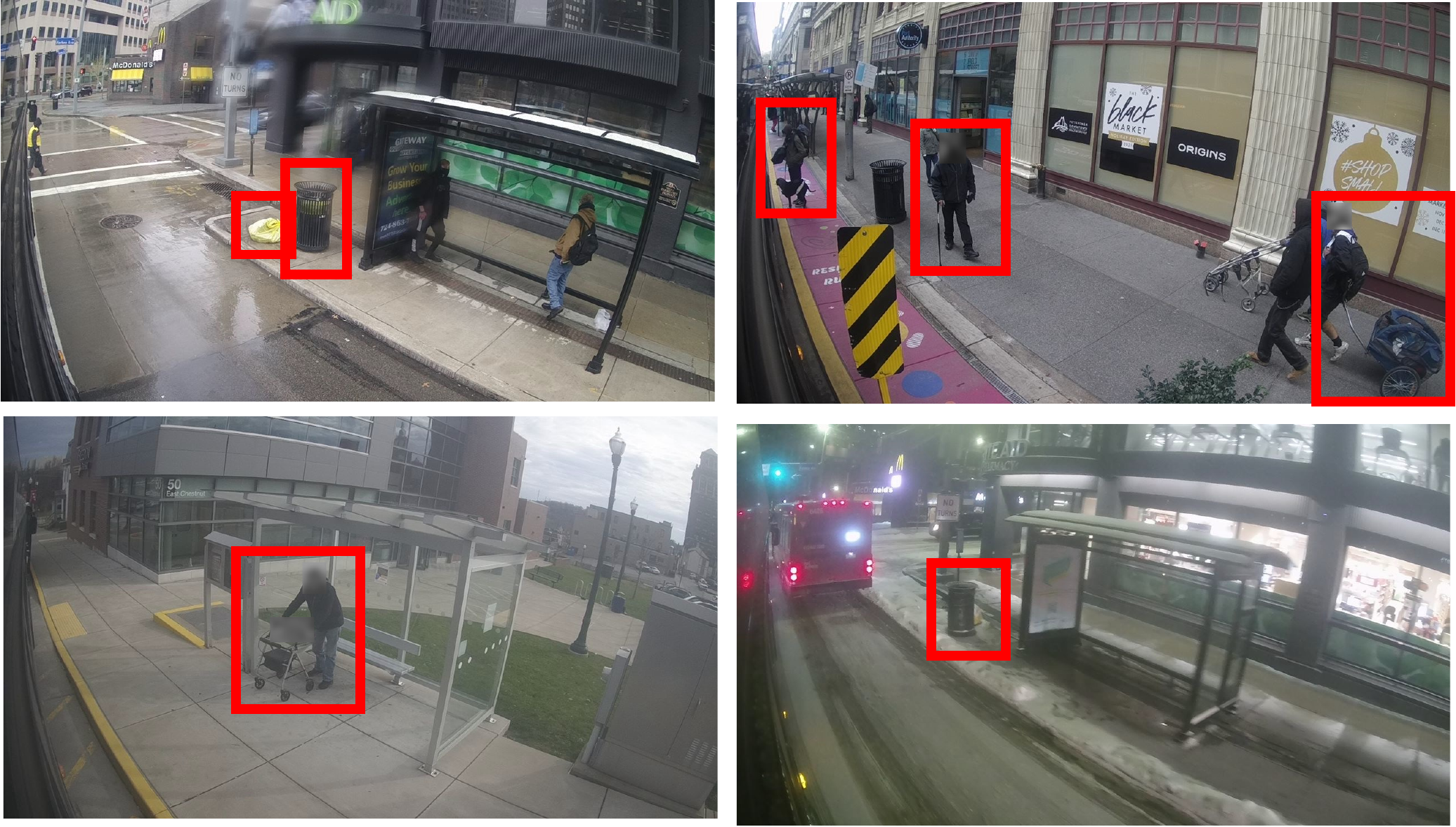}
    \caption{\textbf{Commuter Bus Dataset:} The captured data has a unique viewpoint, the average size of objects is small and captured under a wide variety of lighting conditions. Top-left and bottom-right images depict the same bus-stop and trashcan at different time of day and season.}
    \label{fig:busedge-imagery}
\end{figure}

\noindent
\textbf{Commuter Bus Dataset:} We curated this dataset from a commuter bus running on an urban route. The 720p camera of interest has a unique viewpoint and scene geometry (See Figure~\ref{fig:busedge-imagery}). Annotated categories are trashcans and garbage bags (to help inform public services) and people with special needs (using wheelchairs or strollers), which are a rare occurrence. The dataset size is small with only 750 annotated images (split into 70\% training and 30\% testing). This is an extremely challenging dataset due to it's unique viewpoint, small object size, rare categories, along with variations in lighting and seasons. 

\subsection{Evaluation Details}
\label{sec:considerations}

\noindent
We perform apples-to-apples comparisons on the same detector trained using the datasets with identical training schedule and hardware.

\noindent
\textbf{Data:} We compare to methods that were trained on fixed training data. In contrast, sAP leaderboards~\cite{li2020towards} don't restrict data and evaluate on different hardware. We compare with~\cite{ghosh2021adaptive, li2020towards} from leaderboard, which follow the same protocols. Other methods on the leaderboard use additional training data to train off-the-self detectors. Our detectors would see similar improvements with additional data.

\noindent
\textbf{Real-Time Evaluation:} We evaluate using \textbf{Streaming AP (sAP)} metric proposed by~\cite{li2020towards}, which integrates latency and accuracy into a single metric. Instead of considering models via accuracy-latency tradeoffs~\cite{huang2017speed}, real-time performance can be evaluated by applying real-time constraints on the predictions~\cite{li2020towards}. Frames of the video are observed every 33 milliseconds (30 fps) and predictions for every frame must be emitted \textbf{before} the frame is observed (forecasting is necessary). For a fair comparison, \textbf{sAP} requires evaluation on the same hardware. Streaming results are not directly comparable with other work~\cite{thavamani2021fovea, yang2022streamyolo, li2020towards} as they use other hardware (say, V100 or 2080Ti), thus we run the evaluation on our hardware (Titan X).

\noindent
{\it Additional details are in the Supplementary.}

\begin{figure}[t!]
    \centering
    \includegraphics[width=1\linewidth]{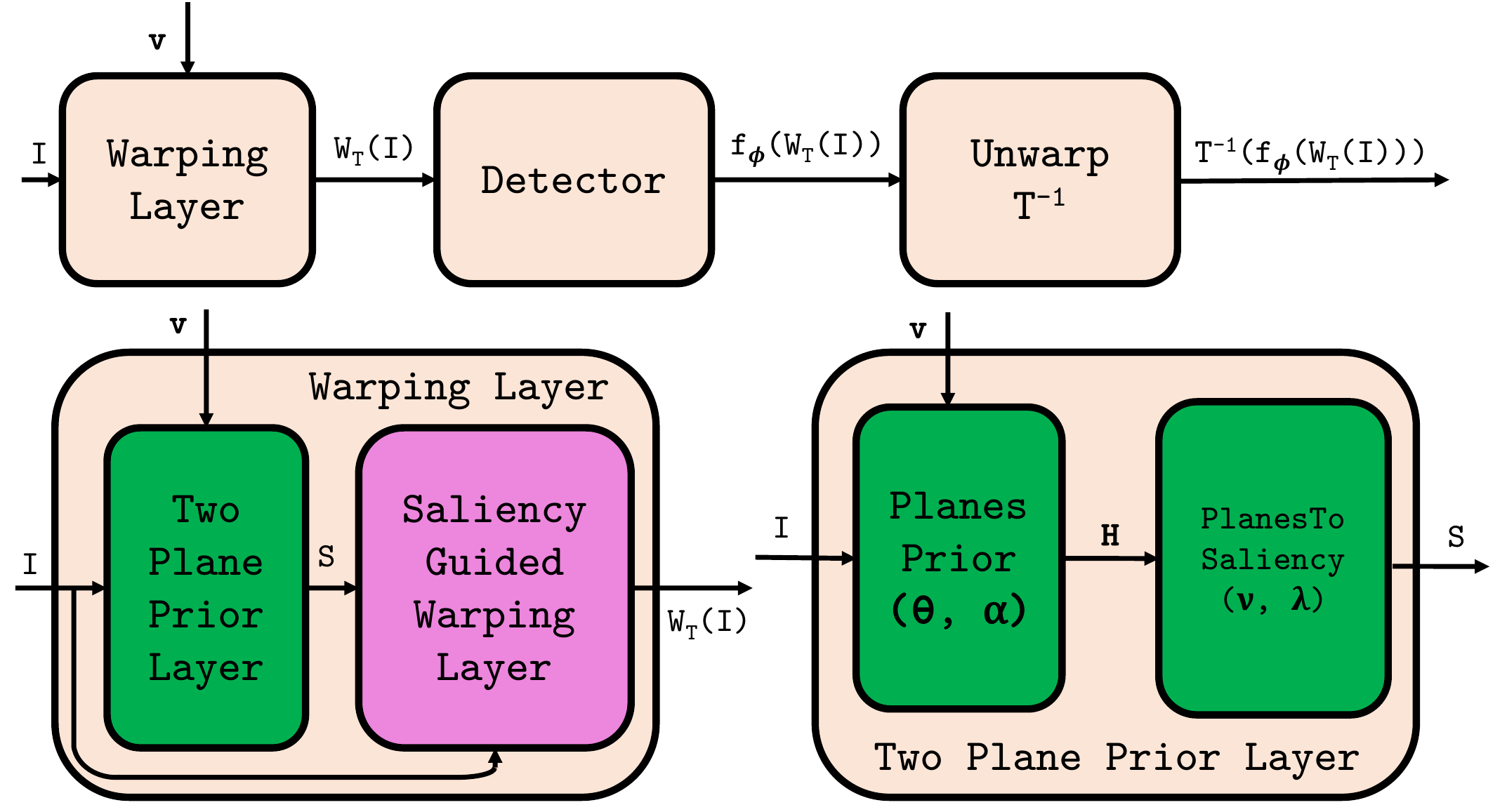}
    \caption{\textbf{Two-Plane Prior as a Neural Layer:} We implemented our approach as a global prior that is learned end-to-end from labelled data. Our prior is dependent on a vanishing point estimate to specify the viewing direction of the camera.}
    \label{fig:implementation}
\end{figure}

\section{Results and Discussions}
\label{sec:results}
\noindent
\textbf{Accuracy-Latency Comparisons:} On Argoverse-HD, we compare with Faster R-CNN with naive downsampling (Baseline) and Faster R-CNN paired with adaptive sampling from Fovea~\cite{thavamani2021fovea} which proposed two priors, a dataset-wide prior ($S_{D}$) and frame-specific temporal priors ($S_{I}$ \& $L: S_{I}$; from previous frame detections). 

\noindent
Two-Plane Prior improves (Table~\ref{table:baselinekiller}) upon baseline at the same scale by \green{+6.6 $AP$} and over SOTA by \green{+2.7 $AP$}. For small objects, the improvements are even more dramatic, our method improves accuracy by \green{+9.6 $AP_{S}$} or \green{195\%} over baseline and \green{+4.2 $AP_{S}$} or \green{45\%} over SOTA respectively. Surprisingly, our method at 0.5x scale improves upon Faster R-CNN at 0.75x scale by \green{+1.6 $AP$} having latency improvement of 35\%. Our Two-Plane prior trained via pseudo-labels comes very close to SOTA which employ ground truth labels, the gap is only \maroon{-1 $AP$} and improves upon Faster R-CNN model trained on ground truth by \green{+2.9 AP} inferred at the same scale.

\begin{table}[t]
{\footnotesize
\centering
\setlength{\tabcolsep}{2.6pt}
\begin{tabular}{lllllll}
\toprule
Method            & Scale   & $AP$ & $AP_S$ & $AP_M$ & $AP_L$ & Latency (ms) \\
\midrule
Faster R-CNN               & 0.5x    & 24.2  & 4.9    & 29.0   & 50.9  & $78.4 \pm 1.8$ \\
\midrule
Fovea ($S_{D}$)~\cite{thavamani2021fovea}   & 0.5x          & 26.7  & 8.2    & 29.7   & 54.1 & $83 \pm 2.5$ \\
Fovea ($S_{I}$)~\cite{thavamani2021fovea}   & 0.5x            & 28.0 & 10.4   & 31.0   & \textbf{54.5} & $85 \pm 2.7$ \\
Fovea (L:$S_{I}$)~\cite{thavamani2021fovea}   & 0.5x           & 28.1    & 10.3   & 30.9   & 54.1  &  $85.4 \pm 2.7$ \\
\midrule
Two-Plane Pr. (Pseudo.) & 0.5x & 27.1   & 9.8    & 28.9   & 50.2 & $104.5 \pm 8.5$ \\
Two-Plane Prior   & 0.5x          & \textbf{30.8} & \textbf{14.5}   & \textbf{31.6}   & 52.9 & $105 \pm 8.5$ \\
\midrule
\multicolumn{7}{c}{Baseline at higher scales} \\
\midrule
Faster R-CNN  & 0.75x         & 29.2 & 11.6   & 32.1   & 53.3 & $142 \pm 2.5$ \\
Faster R-CNN & 1.0x         & 33.3  & 16.8   & 34.8   & 53.6 & $220 \pm 1.7 $ \\
\bottomrule
\end{tabular}
\caption{\textbf{Evaluation on Argoverse-HD:} Two-Plane Prior outperforms both SOTA's dataset-wide and temporal priors in overall accuracy. Our method improves small object detection by \green{$+4.1 AP_{S}$} or 39\% over SOTA.}
\label{table:baselinekiller}
}
\vspace{-0.1in}
\end{table}

\begin{figure}
    \centering
    \includegraphics[width=0.49\linewidth]{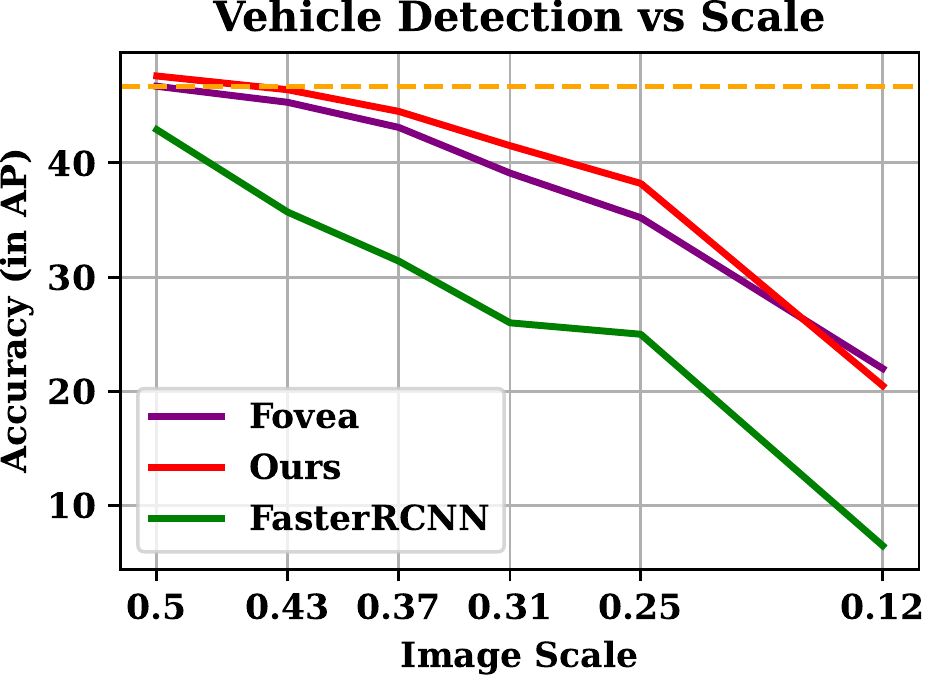}
    \includegraphics[width=0.49\linewidth]{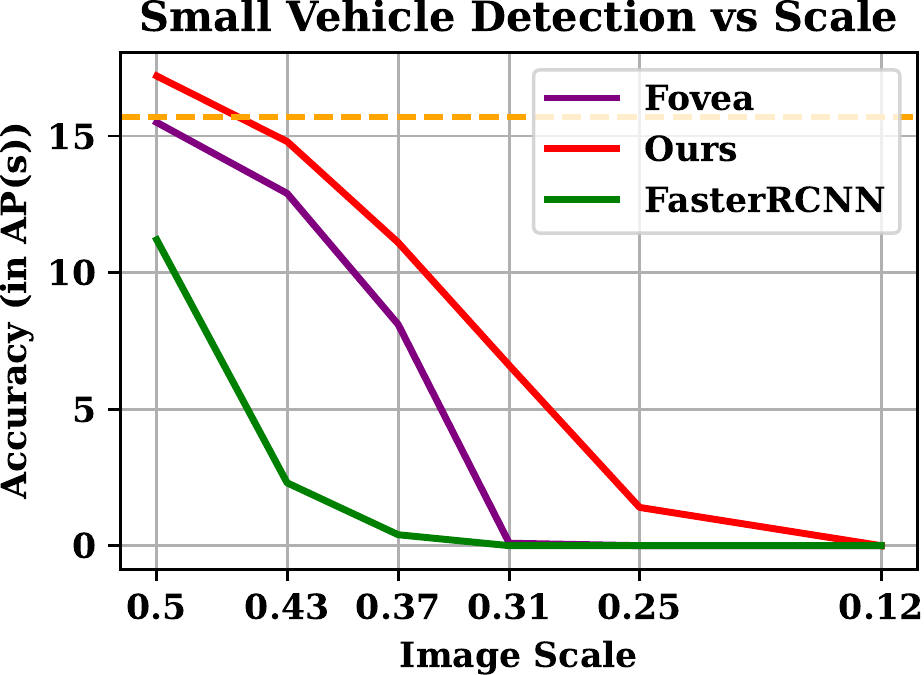}
\caption{\textbf{WALT All-Viewpoints Split:} Our approach (Two-Plane Prior) shows improved overall performance over naive downsampling and state-of-the-art adaptive sampling technique, specially for small objects at all scales (starting from 0.5x). Horizontal line (orange) indicates performance at maximum possible scale (0.6x) the base detector was trained at (memory constraints).}
    \label{fig:walt-scale-fig}
\end{figure}

\noindent
WALT dataset~\cite{reddy2022walt} comprises images (only images with notable changes are saved) and not videos, we compare with Fovea~\cite{thavamani2021fovea} paired with the dataset-wide prior. We observe similar trends on both splits (Table~\ref{table:viewpoint-generalization} and Fig~\ref{fig:walt-scale-fig}) and note large improvements over baseline and a consistent improvement over Fovea~\cite{thavamani2021fovea}, specially for small objects.

\begin{table}[]
{\footnotesize
\centering
\setlength{\tabcolsep}{2.5pt}
\begin{tabular}{lllllll}
\toprule
ID & Method                              & Scale(s) & $sAP$ & $sAP_S$ & $sAP_M$ & $sAP_L$ \\
\midrule
1 & Streamer~\cite{li2020towards}   & 0.5x &  18.3           &  3.1      &  15.0      & 40.9       \\
\midrule
2 & 1 + Adascale~\cite{chin2019adascale} & 0.2x-0.6x & 13.4   & 0.2    & 8.6  & 37.4 \\
3 & Adaptive Streamer~\cite{ghosh2021adaptive}  & 0.2x-0.6x  & 21.3  & 4.2    & 18.8  & 47.0       \\
4 & 1 + FOVEA ($S_I$)~\cite{thavamani2021fovea}     & 0.5x                & 24.1  & 8.4    & 24.7   & 48.7   \\
\midrule
5 & 1 + Ours (Avg VP)             & 0.5x  & \textbf{29.9}   & \textbf{13.7}   & \textbf{31.3}   & \textbf{52.2}   \\
6 & 1 + Ours       & 0.5x  & \textbf{30.0}       &  \textbf{13.7}        &  \textbf{31.5}     &   \textbf{52.2}     \\
\midrule
\midrule
7 & 1 + Ours (VP Oracle)        & 0.5x  & 30.7   & 14.5   & 31.6   & 52.9   \\
\bottomrule
\end{tabular}
\caption{\textbf{Streaming Evaluation on Argoverse-HD:} Ours denotes Two-Plane Prior. Every frame's prediction (streamed at 30FPS) must be emitted \textbf{before} frame is observed~\cite{li2020towards} (via forecasting). All methods evaluated on Titan X GPU. Underlying detector (Faster R-CNN) is constant across approaches, improvements are solely from spatial sampling mechanisms. Notice improved detection of small objects by \green{$+5.3 sAP_{S}$} or 63\% over SOTA.} 
\label{table:streaming-eval}
}
\end{table}

\noindent
\textbf{Real-time/Streaming Comparisons:} We use Argoverse-HD dataset, and compare using the \textbf{sAP} metric (Described in Section~\ref{sec:considerations}). Algorithms may choose any scale and frames as long as real-time latency constraint is satisfied. 

\noindent
All compared methods use Faster R-CNN, and we adopt their reported scales (and other parameters). Streamer~\cite{li2020towards} converts any single frame detector for streaming by scheduling which frames to process and interpolating predictions between processed frames. AdaScale~\cite{chin2019adascale} regresses optimal scale from image features to minimize single-frame latency while Adaptive Streamer~\cite{ghosh2021adaptive} learns scale choice in the streaming setting. Both these methods employ naive-downsampling. State-of-the-art, Fovea~\cite{thavamani2021fovea} employs the temporal prior ($S_{I}$). From Table~\ref{table:streaming-eval}, Two-Plane prior outperforms the above approaches by \green{+16.5 $sAP$}, \green{+8.6 $sAP$} and \green{+5.9 $sAP$} respectively. Comparison with~\cite{chin2019adascale, li2020towards, ghosh2021adaptive} shows the limitations of naive downsampling, even when ``optimal'' scale is chosen. Our geometric prior greatly improves small object detection performance by \green{63\%} or \green{+5.3 $sAP_S$} over SOTA. To consider dependence on accurate Vanishing Point detection (and its overheads), we use NeurVPS~\cite{zhou2019neurvps} as oracle (we simulate accurate prediction with zero delay) to obtain an upper bound, we observe even average vanishing point location's performance is within 0.8 $sAP$.

\noindent
\textbf{Accuracy-Scale Tradeoffs:}  We experiment with WALT \textbf{All-Viewpoints} split to observe accuracy-scale trade-offs. The native resolution ($4K$) of the dataset is extremely large and the gradients don't fit within 12 GB memory of our Titan X GPU, thus we cropped the skies and other static regions to reduce input scale (1x) to $1500 \times 2000$. Still, the highest scale we were able to train our baseline Faster R-CNN model is 0.6x. So, we use aggressive downsampling scales $\{0.5, 0.4375, 0.375, 0.3125, 0.25, 0.125 \}$. The results are presented in Figure~\ref{fig:walt-scale-fig}. We observe a large and consistent improvement over baseline and Fovea~\cite{thavamani2021fovea}, specially for small objects. For instance, considering performance at 0.375x scale, our approach is better than baseline by \green{+13.1 $AP$} and Fovea by \green{+1.4 $AP$} for all objects.

\noindent
For small objects, we observe dramatic improvement, at scales smaller than 0.375x, other approaches are unable to detect any small objects while our approach does so until 0.125x scale, showing that our approach degrades more gracefully. At 0.375x scale, our approach improves upon Faster R-CNN by \green{+10.7 $AP_S$} and Fovea by \green{+3.0 $AP_S$}.

\begin{table}[t]
{\footnotesize
\centering
\setlength{\tabcolsep}{4pt}
\begin{tabular}{lllllll}
\toprule
Method              & Scale   & $AP$ &  $AP_S$ & $AP_M$ & $AP_L$ \\
\midrule
Faster R-CNN & 0.25x & 16.7  & 0	& 7.2 & 42.3 \\
FOVEA ($S_{D}$)~\cite{thavamani2021fovea} & 0.25x & 23.2 &	0	& 13.9 &  54.1 \\
\midrule
Two-Plane Prior & 0.25x & \textbf{25.2}	& \textbf{1.0} &	\textbf{18.2} &	\textbf{54.5} \\
\midrule
\midrule
Faster R-CNN & 0.5x        & 29.2    & 4.9    & 24.7   & 55.5   \\
FOVEA ($S_{D}$)~\cite{thavamani2021fovea} & 0.5x        & 34.4    & 8.7    & 30.5   & \textbf{59.4}   \\
\midrule
Two-Plane Prior & 0.5x  & \textbf{36.4}  & \textbf{11.6}   & \textbf{32.3}   & 59.0   \\
\midrule
\multicolumn{6}{c}{Baseline at higher scales} \\
\midrule
Faster R-CNN & 0.6x       &   33.2      &  9.3      &  28.6      &  56.7  \\   
Faster R-CNN\textbf{*} & 1.0x       &   34.3         &  12.6      &  30.7      &  54.3     \\
\bottomrule
\end{tabular}
\caption{\textbf{WALT Camera-Split:} The viewpoints on the test set were not seen, and Two-Plane Prior shows better performance over both naive downsampling and state-of-the-art adaptive sampling as it generalizes better to unseen scenes and viewpoints. \textit{*Not trained at that scale due to memory constraints on Titan X.}}
\label{table:viewpoint-generalization}
}
\vspace{-0.2in}
\end{table}

\noindent
\textbf{Generalization to new viewpoints:} We use WALT \textbf{Camera-Split}, the test scenes and viewpoint are unseen in training. E.g., the vanishing point of one of the held-out cameras is beyond the image's field of view. We operate on the same scale factors in the earlier experiments, and results are presented in Table~\ref{table:viewpoint-generalization}. We note lower overall performance levels due to scene/viewpoint novelty in the test sets. Our approach generalizes better due to the explicit modelling of the viewpoint via the vanishing point (See Section~\ref{sec:3dreasoning}). We note trends similar to previous experiment, we demonstrate improvements of \green{+8.5 $AP$} over naive-downsampling and \green{+2.0 $AP$} over Fovea~\cite{thavamani2021fovea} at 0.25x scale. 

\noindent
\textbf{Tracking Improvements}: We follow tracking-by-detection and pair IOUTracker~\cite{bochinski2017high} with detectors on Argoverse-HD dataset. MOTA and MOTP evaluate overall tracking performance. From Table~\ref{table:tracking-improvements}, Our method improves over baseline by \green{+4.8\%} and \green{+0.7\%}. We also focus on tracking quality metrics, Mostly Tracked \% (\textbf{MT\%}) evaluates the percentage of objects tracked for atleast 80\% of their lifespan while Mostly Lost \% (\textbf{ML\%}) evaluates percentage of objects for less than 20\% of their lifespan. In both these cases, our approach improves upon the baseline by \green{+7.6\%} and \green{-6.7\%} respectively. To  autonomous navigation, we define two relevant metrics, namely, average lifespan extension \textbf{(ALE)} and minimum object size tracked \textbf{(MOS)}, whose motivation and definitions can be viewed in supplementary.  We observe that our improvements are better than both Fovea ($S_{D}$) and ($S_{I}$). As we observe, our method improves tracking lifespan and also helps track smaller objects.

\noindent
\textbf{Efficient City-Scale Sensing:} We detect objects on Commuter Bus equipped with a Jetson AGX. Identifying (Recall) relevant frames is key on the edge. Recall of Faster R-CNN with at 1x scale is $43.3 AR$ ($16.9 AR_{S}$) (Latency: 350ms; infeasible for real-time execution) but drops to $31.7 AR$ ($0.5 AR_{S}$) at 0.5x scale when naively down-sampled (Latency: 154ms). Whereas our approach at 0.5x scale improves recall by 42\% over full resolution execution to \green{61.7 $AR$} (\green{16.4 $AR_{S}$}) with latency of 158 ms (+4 ms). 

\noindent
In supplementary, we perform additional comparisons, provide details on handling multiple vanishing points, more tracking results and edge sensing results. We also present some qualitative results and comparisons.

\begin{table}[]
{\footnotesize
\centering
\setlength{\tabcolsep}{2.5pt}
\begin{tabular}{lllllll}
\toprule
Method                              & MOTA $\uparrow$ & MOTP $\uparrow$ & MT\% $\uparrow$ & ML\% $\downarrow$ & MOS $\downarrow$ & ALE\% $\uparrow$ \\
\midrule
Faster RCNN & 39.8 & 82.3 & 30.7 & 35.6 & 37.1 & 59.3 \% \\
Fovea ($S_{D}$)~\cite{thavamani2021fovea} & 43.9 & 81.9 & 34.1 & 31.9 & 34.8 & +5.4 \% \\
Fovea ($S_{I}$)~\cite{thavamani2021fovea} & 44.3 & 81.8 & 36.7 & \textbf{28.4} & 33.8 & +8.4 \% \\
\midrule
 Two-Plane Prior      & \textbf{44.6}       &  \textbf{83.0}        &  \textbf{38.3}     &   28.9  & \textbf{31.6} & \textbf{+9.9 \%} \\
\bottomrule
\end{tabular}
\caption{{\textbf{Tracking Improvements}:} We setup a tracking by detection pipeline and replace the underlying detection method and observe improvements if any. All the detectors employ the Faster R-CNN architecture and are executed at 0.5x scale. We observe improvements in tracking metrics due to Two-Plane Prior.}
\label{table:tracking-improvements}
}
\end{table}

\subsection{Ablation Studies}

\noindent
We discuss some of the considerations of our approach through experiments on the Argoverse-HD dataset.

\noindent
\textbf{Ground Plane vs Two-Plane Prior:} We discussed the rationale of employing multiple planes in Fig~\ref{fig:saliency-explanation}, and our results are consistent. From Table~\ref{table:design-considerations}, Two-Plane Prior outperforms Ground Plane prior considerably (\green{+1.8 $AP$}). Ground Plane Prior outperforms Two-Plane Prior on small objects by \green{+1 $AP_S$} but is heavily penalized on medium (\maroon{-4.1 $AP_M$}) and large objects (\maroon{-4.6 $AP_L$}). This is attributed to heavy distortion of tall and nearby objects, and objects that are not on the plane (Figure~\ref{fig:saliency-explanation}). Lastly, this prior was difficult to learn, the parameter space severely distorted the images (we tuned initialization and learning rate). Thus we did not consider this prior further. The second plane acts as a counter-balance and that warping space is learnable.

\noindent
\textbf{Vanishing Point Estimate Dependence:} From Table~\ref{table:design-considerations}, dominant vanishing point in autonomous navigation is highly local in nature, and estimating VP improves the result by \green{+1.2 AP}. Estimating the vanishing point is a design choice, it's important for safety critical applications like autonomous navigation (performance while navigating turns) however might be omitted for sensing applications.

\begin{table}[t]
{\footnotesize
\centering
\setlength{\tabcolsep}{4pt}
\begin{tabular}{lllllll}
\toprule
Method              & Scale   & $AP$ &  $AP_S$ & $AP_M$ & $AP_L$ \\
\midrule
Ground Plane Prior  & 0.5x & 29.1  & 15.5 & 27.5 & 48.3 \\
Two-Plane Prior (Psuedo.) & 0.5x & 27.1 & 9.8 & 28.9 & 50.2 \\
Two-Plane Prior (Avg VP) & 0.5x & 29.6 & 12.7 & 30.7 & 52.7 \\
Two-Plane Prior & 0.5x  & 30.8 & 14.5 & 31.6 & 52.9 \\
\bottomrule
\end{tabular}
\caption{\textbf{Ablation Study on Argoverse-HD} to justify our design choice of using two planes, dependence on accurate vanishing point detection and choice of pseudo labels vs ground truth.}
\label{table:design-considerations}
}
\vspace{-0.2in}
\end{table}

\noindent
\textbf{Using Pseudo Labels vs Ground Truth:} Table~\ref{table:design-considerations} shows there is still considerable gap (\maroon{-3.7 $AP$}) between the Two-Plane Prior trained from pseudo labels and ground truth. We observe that the model under-performs on $stopsign$, $bike$ and $truck$ classes, which are under-represented in the COCO dataset~\cite{lin2014microsoft} compared to $person$ and $car$ classes. Performance of the pre-trained model on these classes is low even at 1x scale. Hence, we believe that the performance difference is an artifact of this domain gap.

% \section{Efficient City-Scale Sensing}
% \label{sec:application}
% \input{cvpr2023/busedge.tex}

\section{Conclusions}
\label{sec:conclusion}
\noindent
In this work, we proposed a learned two-plane perspective prior which incorporates rough geometric constraints from 3D scene interpretations of 2D images to improve object detection. We demonstrated that (a) Geometrically defined spatial sampling prior significantly improves detection performance over multiple axes (accuracy, latency and memory) in terms of both single-frame accuracy and accuracy with real-time constraints over other methods. (b) Not only is our approach is more accurate when adaptively down-sampling at all scales, it degrades much more gracefully for small objects, resulting in latency and memory savings. (c) As our prior is learned end-to-end, we can improve a detector's performance at lower scales for ``free''. (d) Our approach generalizes better to new camera viewpoints and enables efficient city-scale sensing applications. Vanishing point estimation is the bottleneck of our approach~\cite{coughlan2000manhattan, zhou2019neurvps, lin2022deep, liu2021vapid} for general scenes, and increasing efficiency of its computation we will see substantial improvements. Investigating geometric constraints to improve other aspects of real-time perception systems as future work, like object tracking and trajectory understanding and forecasting, is promising. 

\noindent 
\textbf{Societal Impact} Our approach has strong implications for autonomous-driving and city-scale sensing for smart city applications, wherein efficient data processing would lead to more data-driven decision-making and public policies. However, privacy is a concern, and we shall release the datasets after anonymizing people and license plates.

\noindent 
\textbf{Acknowledgements:} This work was supported in part by an NSF CPS Grant CNS-2038612, a DOT RITA Mobility-21 Grant 69A3551747111 and by General Motors Israel.

%%%%%%%%% REFERENCES

% \balancecolsandclearpage

{\small
\bibliographystyle{ieee_fullname}
\bibliography{egbib}
}

\appendix
\section{Implementation, Training and Evaluation Details}

\subsection{Implementation Details}

\noindent
We implemented our approach using Pytorch~\cite{pytorch} and mmdetection~\cite{mmdetection}. The two-plane perspective prior is implemented as a neural network layer with learnable parameters that are global (fixed warps parameterized by vanishing point; Figure~\ref{fig:implementation}). We employ differentiable versions of Direct Linear Transform and $warp\_perspective$ from Kornia~\cite{kornia}, while we reuse implementation of separable neural warps from~\cite{thavamani2021fovea}. To detect vanishing points, we employ NeurVPS~\cite{zhou2019neurvps} for fixed cameras. We use VPNet~\cite{liu2020d} with ResNet18 backbone for autonomous navigation.

\noindent
\textbf{Parameter Initialization} We selected one representative image, and initialized the learnable parameters (i.e. $\theta$'s and $\alpha$'s) via visual inspection. The guiding principal was to enlarge far objects while trying to distort the close-by objects as less as possible. The same initial parameters are used for all the datasets. Please look at our code for the initial parameters.

\subsection{Evaluation Details}

\noindent
\textbf{Detection Model Choice} We experiment with Faster R-CNN as our base detection model as prior work has shown it occupies the optimal sweet-spot~\cite{li2020towards} w.r.t latency and accuracy on modern GPUs. However, our approach is agnostic to the choice of detector and our results generalize to other detectors.

\noindent
\textbf{Latency:} We capture end-to-end latency in milliseconds, that includes image pre-processing, network inference and post-processing, following protocol from prior work~\cite{thavamani2021fovea}.

\noindent
\textbf{Scale} The image down-sampling factor is equal in both spatial dimensions. So an image originally $1920\times1200$ (1x scale) when down-sampled to 0.25x is a $480\times300$ image.

\subsection{Training Details}

\noindent
For training our proposed approaches, to train the Faster R-CNN model we use the Adam optimizer with a learning rate of $3\times10^{-4}$. For training any methods by Fovea~\cite{thavamani2021fovea} we follow their protocol. We follow the protocol mentioned by~\cite{ghosh2021adaptive, chin2019adascale} for training their approach. 

\noindent
\textbf{Argoverse-HD} We considered the same base architecture (Faster R-CNN) for all the methods. We compare with SOTA~\cite{thavamani2021fovea} using models provided by their public code release, and follow the training protocol prescribed in their work, training our models for 3 epochs.

\noindent
\textbf{WALT} We considered the same base architecture (Faster R-CNN) for all the methods. We trained the models (and learnt the warping function parameters, if applicable) using the Adam optimizer with the same learning rate and other parameters for 6 epochs.

\begin{figure*}
    \centering
    \includegraphics[width=\linewidth]{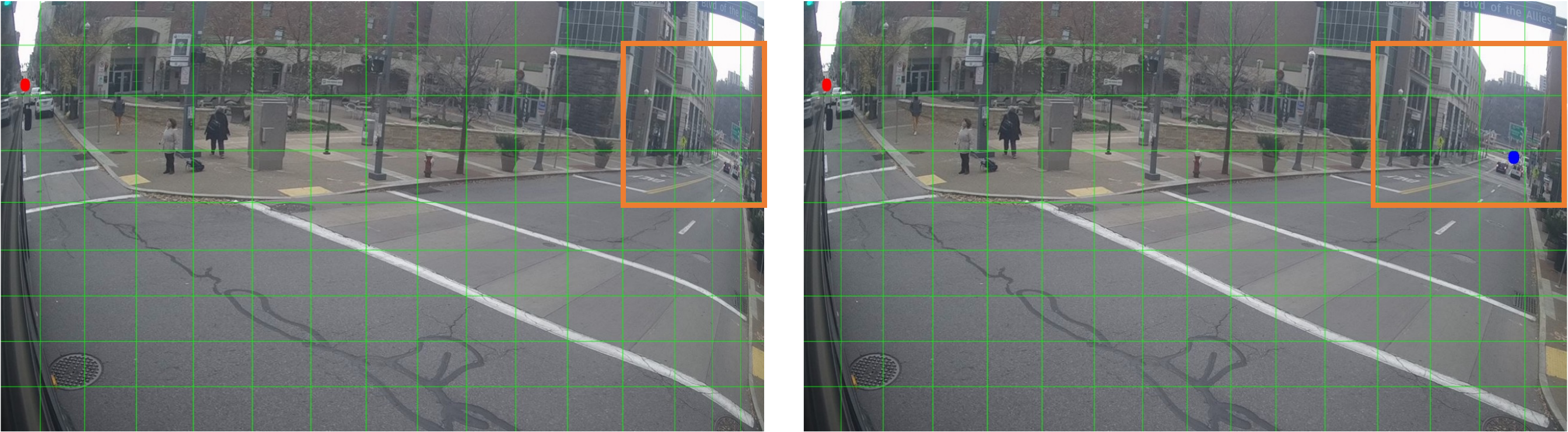}    
    \caption{\textbf{Multiple Vanishing Points:} Saliency from \red{First Vanishing Point} parallel to sidewalk ``compresses'' far away cars on perpendicular road. \blue{Second Vanishing Point} ensures those cars are not compressed. Please zoom in to observe the vanishing points and deformation.}
    \label{fig:multivps}
\end{figure*}

\noindent
\textbf{Vanishing Point Estimation} For NeurVPS, we directly employ the pre-trained model trained on Natural Scenes (TMM17) dataset~\cite{zhou2017detecting} part of their public code release. While for VPNet~\cite{liu2020d}, as there is no public code release, we implement this architecture employing a ResNet18 backbone attached to a modified YOLO head. We omit the upsampling  refinement procedure described in~\cite{liu2020d}, as model's median error in vanishing point prediction is around 10 pixels with an average latency of 28 ms, which is sufficient for our method to work. The off-the-shelf model is executed at $n_{v} = 30$ to amortize the cost of executing this model. We also tried using LaneAF~\cite{abualsaud2021laneaf} to obtain lane lines (similar latency), however, we observed the method was prone to errors while clustering lines and obtaining the vanishing point.

\section{Multiple Vanishing Points}

\noindent Our method can consider additional planes that correspond to lines meeting at a different vanishing point. For example, a traffic camera with a wide field of view that is placed at an intersection observing two roads simultaneously would benefit from this. Assuming $N$ vanishing points, considering Saliency $S_{v_{i}}$ corresponding to vanishing point $v_{i}$,

\begin{equation}
    S = \sum_{i = 0}^{N} \lambda_{i} S_{v_{i}}
\end{equation}

where $\lambda_{i}$'s are learnable, initialized as $\frac{1}{N}$. Please observe the case of $N=2$ in an image from the commuter bus dataset in Figure~\ref{fig:multivps}, wherein combining saliencies from two vanishing points (obtained from~\cite{lin2022deep}) ensures far away objects of interest are sampled more. 

We observed in the datasets we considered, multiple vanishing points were rare as it generally requires a camera with a large field of view. Thus, we employed models~\cite{zhou2019neurvps, liu2020d} trained on Natural Scenes dataset, which predict only one vanishing point. However, vanishing points can be estimated from other methods~\cite{zhou2019neurvps, lin2022deep, liu2021vapid} which do predict all the vanishing points, but incur higher overheads.

\section{Results on Another Detector}

\noindent
Adaptive spatial sampling mechanisms leverage and exploit priors corresponding to the input images in a way that is agnostic to the detection method. We expect our approach to generalize across detectors, similar to observations by such warping mechanisms and saliency priors proposed earlier~\cite{thavamani2021fovea}. We choose RetinaNet~\cite{lin2017focal}, a popular single-stage object detector as our archetypal example (Faster R-CNN~\cite{ren2015faster} is the two-stage archetype). Results can be viewed in Table~\ref{table:appendix-alternate-detector}. Our approach improves upon both the baseline Faster R-CNN and SOTA, specially for small and medium sized objects, following the trends observed in the main manuscript.

\begin{table}[]
{\footnotesize
\centering
\setlength{\tabcolsep}{3pt}
\begin{tabular}{llllllll}
\toprule
Method                                    & Scale & $AP$ & $AP_S$ & $AP_M$ & $AP_L$ \\
\midrule
RetinaNet & 0.5x & 22.6 & 4.0 & 22.0 & 53.1 \\
Fovea ($S_{I}$)~\cite{thavamani2021fovea}                               & 0.5x & 24.9 & 7.1 & 27.7 & 50.6 \\
Two-Plane Prior & 0.5x & \textbf{26.3} & \textbf{10.1} & \textbf{29.2} & 50.5 \\
\midrule
\multicolumn{6}{c}{Baseline at higher scales} \\
\midrule
RetinaNet & 0.75x & 29.9 & 9.7 & 32.5 & 54.2 \\
\bottomrule
\end{tabular}
\caption{\textbf{Alternate Detector:} We replace Faster R-CNN with RetinaNet (archetypal one-stage detector), and observe considerable improvements over Baseline (RetinaNet with uniform downsampling) and SOTA trained on Argoverse-HD dataset.} 
\label{table:appendix-alternate-detector}
}
\end{table}

\begin{table}[]
{\footnotesize
\centering
\setlength{\tabcolsep}{3pt}
\begin{tabular}{llllllll}
\toprule
Method                                    & Scale & Model & $AP$ & $AP_S$ & $AP_M$ & $AP_L$ \\
\midrule
Faster R-CNN                               & 0.5x & COCO  & 15.3 & 1.1 & 12.5 & 40.5 \\
Faster R-CNN                               & 0.5x & AVHD  & 15.1 & 1.0 & 10.6 & 39.0 \\
\midrule
Fovea ($S_{D}$)~\cite{thavamani2021fovea}  & 0.5x & AVHD  & 13.7 & 1.3 & 10.0 & 34.7 \\
Fovea ($S_{I}$)~\cite{thavamani2021fovea}  & 0.5x & AVHD  & 16.4 & 2.1 & 12.8 & 38.6 \\
% \midrule
% Fovea ($S_{D}$)~\cite{thavamani2021fovea} (BDD Sal.)  & 0.5x & AVHD &  & & &  \\ 
\midrule
Two-Plane Prior (Psuedo.)                  & 0.5x & AVHD  & 16.2 & \textbf{4.7} & \textbf{15.9} & 33.3 \\
Two-Plane Prior (Psuedo.)                  & 0.5x & COCO  & \textbf{20.9} & \textbf{5.8} & \textbf{19.4} & \textbf{44.2} \\
\midrule
\multicolumn{7}{c}{Baseline at higher scales} \\
\midrule
Faster R-CNN  & 0.75x                      & AVHD & 19.7 & 3.0 & 16.1 & 44.2 \\
Faster R-CNN & 0.75x                       & COCO  & 20.3 & 3.7 & 18.2 & 45.3 \\
Faster R-CNN  & 1x                         & AVHD & 22.6 & 5.7 & 20.1 & 45.7 \\
Faster R-CNN  & 1x                         & COCO  & 23.1 & 6.5 & 21.7 & 46.1 \\

\bottomrule
\end{tabular}
\caption{\textbf{Generalization to BDD100K:} Scale in this case is fixed to 0.5x, AVHD refers to Argoverse-HD and COCO datasets respectively. AVHD models are finetuned from the pre-trained COCO model. We compare generalization on the BDD100K dataset. Our method assumes availability of training set images of BDD100K \textbf{and not labels}, we generate pseudo-labels from the available model (Section 3.6) to learn the Two Plane prior.} 
\label{table:appendix-generalization-BDD100K}
}
\end{table}

% \section{Results with another backbone}

% \noindent
% \blue{Changing the backbone makes things more accurate.}

% \section{Some Modern Detectors}

% \noindent
% \blue{The method also improves accuracy of modern heavy detectors.}

\begin{table}[]
{\footnotesize
\centering
\setlength{\tabcolsep}{2.5pt}
\begin{tabular}{lllll}
\toprule
Method                              & $sAP$ & $sAP_S$ & $sAP_M$ & $sAP_L$ \\
\midrule
StreamYOLO-L~\cite{yang2022streamyolo} & 25.9 & 8.6 & 24.2 & 40.9 \\
StreamYOLO-M~\cite{yang2022streamyolo} & 25.9 & 9.2 & 24.8 & 41.0 \\
StreamYOLO-S~\cite{yang2022streamyolo} & 29.6 & 11.0 & 30.9 & 51.6 \\
\midrule
 Ours      & \textbf{30.0}       &  \textbf{13.7}        &  \textbf{31.5}     &   \textbf{52.2}     \\
\bottomrule
\end{tabular}
\caption{{\textbf{``Real-Time'' Detectors}:} Streaming Comparison on Argoverse-HD on Titan X. StreamYOLO-M and StreamYOLO-L single-frame latency is \red{45.8 ms} and \red{62.9 ms} respectively, is greater than 33ms, violating~\cite{yang2022streamyolo}'s ``real-time'' restriction. StreamYOLO-S satisfies (\green{20.8 ms}), hence has better performance.}
\label{table:appendix-streamyolo}
}
\end{table}

\section{Additional Results on Autonomous Driving}

\begin{table*}[t!]
{\footnotesize
\centering
\setlength{\tabcolsep}{3.5pt}
\begin{tabular}{lllllllllllllllll}
\toprule
& Method            & Scale   & $AP$   & $AP_{50}$ & $AP_{75}$ & $AP_S$ & $AP_M$ & $AP_L$ & person & mbike & tffclight & bike & bus  & stop & car  & truck \\
\midrule
& Faster R-CNN (Pre.) & 0.5x    & 21.5 & 35.8    & 22.3    & 2.8 & 22.4 & 50.6 & 20.8 & 9.1 & 13.9 & 7.1 & 48.0 & 16.1 & 37.2 & 20.2 \\
& Faster R-CNN               & 0.5x    & 24.2 & 38.9    & 26.1    & 4.9    & 29.0   & 50.9   & 22.8   & 7.5   & 23.3      & 5.9  & 44.6 & 19.3 & 43.7 & 26.6 \\
\midrule
% Fovea ($S_{D}$)~\cite{thavamani2021fovea} (Pre.)   & 0.5x &  23.3 & 40.0 & 22.9 & 5.4 & 25.5 & 48.9 & 20.9 & 13.7 & 12.2 & 9.3 & 50.6 & 20.1 & 40.0 & 19.5 \\
% Fovea ($S_{I}$)~\cite{thavamani2021fovea} (Pre.)  & 0.5x  & 24.1 & 40.7 & 24.3 & 8.5 & 24.5 & 48.3 & 23.0 & 17.7 & 15.1 & 10.0 & 49.5 & 17.5 & 41.0 & 19.4 \\
% Fovea (L:$S_{I}$)~\cite{thavamani2021fovea} (Pre.) & 0.5x & 24.0 & 40.5 & 24.3 & 7.4 & 26.0 & 48.2 & 22.5 & 14.9 & 14.0 & 9.5 & 49.7 & 20.6 & 41.0 & 19.9 \\
& Fovea (Learned Nonsep.)~\cite{thavamani2021fovea} & 0.5x & 25.9  & 42.9  & 26.5  & 10.0  & 28.4 & 48.5  & 25.2  & 11.9  & 20.9  & 7.1  & 39.5  & 25.1 & 49.4  & 28.1 \\
& Fovea (Learned Sep.)~\cite{thavamani2021fovea} & 0.5x & 27.2 & 44.8 & 28.3 & 12.2 & 29.1 & 46.6 & 24.2 & 14.0 & 22.6 & 7.7 & 39.5 & 31.8 & 50.0 & 27.8 \\
\midrule
\parbox[t]{2mm}{\multirow{3}{*}{\rotatebox[origin=c]{90}{SOTA}}} & Fovea ($S_{D}$)~\cite{thavamani2021fovea}   & 0.5x          & 26.7 & 43.3    & 27.8    & 8.2    & 29.7   & 54.1   & 25.4   & 13.5  & 22.0      & 8.0  & 45.9 & 21.3 & 48.1 & 29.3 \\
& Fovea ($S_{I}$)~\cite{thavamani2021fovea}   & 0.5x            & 28.0 & 45.5    & 29.2    & 10.4   & 31.0   & \textbf{54.5}   & 27.3   & 16.9  & \textbf{24.3}      & 9.0  & 44.5 & 23.2 & 50.5 & 28.4 \\
& Fovea (L:$S_{I}$)~\cite{thavamani2021fovea}   & 0.5x           & 28.1 & 45.9    & 28.9    & 10.3   & 30.9   & 54.1   & 27.5   & \textbf{17.9}  & 23.6      & 8.1  & 45.4 & 23.1 & 50.2 & 28.7 \\
\midrule
\parbox[t]{2mm}{\multirow{4}{*}{\rotatebox[origin=c]{90}{Ours}}}
 & Ground Plane Prior  & 0.5x       & \textbf{29.1} & \textbf{46.2}    & \textbf{30.5}    & \textbf{15.5}   & 27.5   & 48.3   & \textbf{28.6}   & 15.0  & 10.4      & \textbf{10.5} & 33.7 & \textbf{46.2} & \textbf{55.4} & \textbf{32.9} \\
& Two-Plane Pr. (Psuedo.) & 0.5x & 27.1 & 43.4    & 28.4    & 9.8    & 28.9   & 50.2   & \textbf{28.2}   & 16.2  & 20.6      & 8.1  & \textbf{50.8} & 26.1 & 45.2 & 21.7 \\
& Two-Plane Pr. (Avg VP) & 0.5x & \textbf{29.6} & \textbf{45.9}    & \textbf{31.6}    & \textbf{12.7}   & \textbf{30.7}   & 52.7   & \textbf{28.4}   & 14.3  & \textbf{24.3}      & \textbf{11.9} & 38.5 & \textbf{31.6} & \textbf{53.9} & \textbf{34.2} \\
& Two-Plane Prior   & 0.5x          & \textbf{30.8} & \textbf{47.2}    & \textbf{33.2}    & \textbf{14.5}   & \textbf{31.6}   & 52.9   & \textbf{30.0}   & 16.7  & 24.1      & \textbf{13.7} & 35.9 & \textbf{35.9} & \textbf{55.3} & \textbf{35.3} \\
\midrule
\multicolumn{17}{c}{Baseline at higher scales} \\
\midrule
& Faster R-CNN  & 0.75x         & 29.2 & 47.6    & 31.1    & 11.6   & 32.1   & 53.3   & 29.6   & 12.7  & 30.8      & 7.9  & 44.1 & 29.8 & 48.8 & 30.1 \\
& Faster R-CNN & 1.0x         & 33.3 & 53.9    & 35.0    & 16.8   & 34.8   & 53.6   & 33.1   & 20.9  & 38.7      & 6.7  & 44.7 & 36.7 & 52.7 & 32.7 \\
\bottomrule
\end{tabular}
\caption{\textbf{Evaluation on the Argoverse-HD dataset:} This is an expanded version of the Table present in the manuscript. We see improvements for most of the objects that are on the ground, with much better overall performance for both small and medium sized objects along with improvements in $AP_{50}$ and $AP_{75}$. Notice that Two-Plane prior performs at par with SOTA on ``traffic-light" category. Pre. denotes Pretrained model, while Psuedo. denotes model trained with Psuedo-Labels from pretrained model (no access to Argoverse-HD labels) as described in Section 3.6.  \textit{We have bolded all of our method variations that perform better than SOTA.} }
\label{table:appendix-baselinekiller}
}
\end{table*}

\noindent We shall consider the comparisons made in Section 5 on Argoverse-HD in the main manuscript and present some additional results (specially across different categories present in the dataset).

\noindent
\textbf{Comparison with Learned Fovea~\cite{thavamani2021fovea}:} Fovea~\cite{thavamani2021fovea} also proposed end-to-end global, dataset-wide saliency map $S$ learned via backpropagation (Learned Seperable and Learned Nonseperable). However, they observed worse performance compared to their bounding box priors ($S_{D}$ and $S_{I}$), see Table~\ref{table:appendix-baselinekiller}. We show that end-to-end learned saliency is better, with careful geometric parameterization.

\noindent
\textbf{Improved Performance on ground plane:} We observe improved performance over state-of-the-art on every object category for objects on the ground plane (person, traffic light, bike, stop-sign, car, truck) apart from motorbike (See Table~\ref{table:appendix-baselinekiller}). On further observation, this might be an artifact of the label skew of the Argoverse-HD dataset ($mbike$ has the least number of instances).  For objects not on the ground plane, like traffic-light, we observe performance as good as SOTA.

\noindent
\textbf{Generalization to BDD100K:} We compare generalization from approaches trained on Argoverse-HD and COCO datasets to BDD100K MOT dataset (See Table~\ref{table:appendix-generalization-BDD100K}). Our approach assumes access to training set images from BDD100K dataset, but not it’s ground truth labels. As our priors can be learnt without access to ground truth data, we employ the method detailed in Section 3.6 to generate pseudo labels to learn and adapt the geometric parameters on this dataset by training on these pseudo-labels for 1 epoch only. 

We observe that our method nearly matches SOTA when adapted starting from a model finetuned on Argoverse-HD, however, dramatically exceeds it’s performance when adapted from a model solely trained on COCO. We believe the reason for this mismatch is due to catastrophic forgetting~\cite{kemker2018measuring} observed in finetuned models when evaluated on out-of-distribution data. Lastly, the results indicate the benefits of learnability of our perspective prior, we observe increase in performance for ``free’’ even when images are available without access to ground truth.

\noindent 
\textbf{Comparison with ``Real-Time'' Detectors:} Real-time detectors like~\cite{yang2022streamyolo} have been recently proposed which predict boxes $G_{t+1}$ at time $t$ (of frame $F_{t+1}$; available solely during training and not testing) given $F_{t}$ to satisfy sAP. Approaches like these constraint the detector to perform the computation within a latency budget ($<33$ms or 30 FPS). Our methods are complementary to such detectors, as long as \textit{their} constraint is satisfied.

However, real-time detectors (termed as ``fast'' strategy) might be suboptimal~\cite{li2020towards}. Satisfying the real-time detector constraint may not be optimal for every hardware platform, specially on slower edge devices. Such methods~\cite{yang2022streamyolo} are \textbf{not} hardware-agnostic, and model architecture choices are optimized for specific hardware (in their case, for a V100 GPU). On Titan X (See Table~\ref{table:appendix-streamyolo}), their streaming performance (which is hardware dependent) is worse.

\section{Tracking Smaller Objects for Longer}

\noindent We provide an analysis of our approach observing how it improves object tracking. We wish to observe if the gains from our method translates to detecting far-away objects for longer period of time. We employ Argoverse-HD dataset for our experiments which have ground truth object IDs.

\noindent
\textbf{Setup:} We employ a Faster R-CNN as our baseline and the tracker is fixed to IOU Tracker~\cite{bochinski2017high}. We additionally pair the priors proposed by Fovea~\cite{thavamani2021fovea} for comparison. All the detectors are executed at 0.5x scale for fair comparison.

\begin{figure*}
    \centering
    \includegraphics[width=0.49\linewidth]{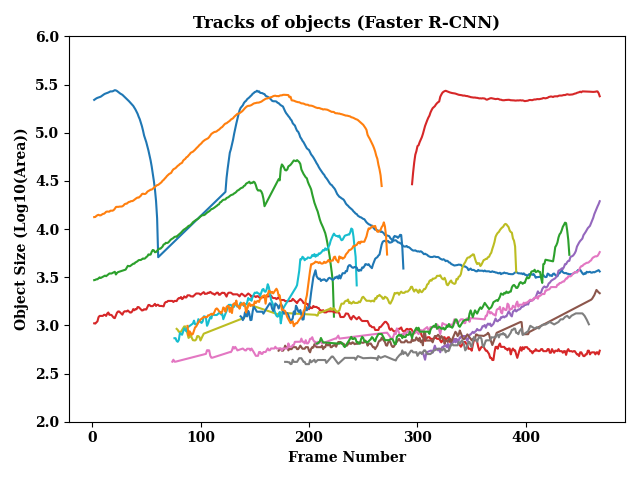}
    \includegraphics[width=0.49\linewidth]{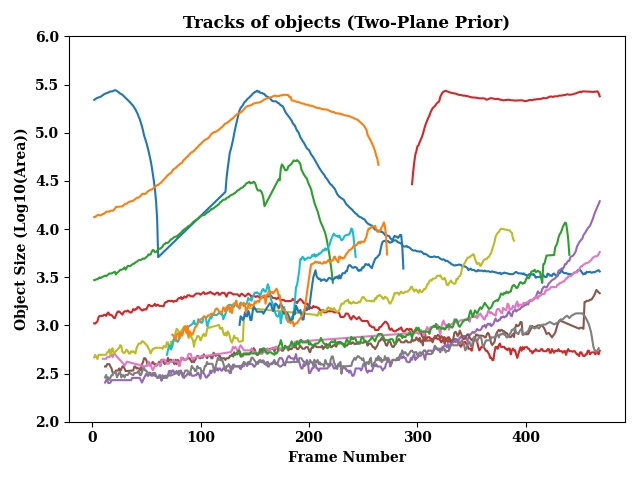}
\caption{\textbf{Tracking Visualization:} To visualize the impact of our two-plane prior, we visualize tracks of length greater than 150 frames tracked by both the methods for a given sequence. We plot object size w.r.t frame numbers (which denotes length). We can observe that some objects are detected earlier and are tracked for a longer time.}
    \label{fig:appendix-track-viz-1}
\end{figure*}

\begin{figure*}
    \centering
    \includegraphics[width=0.49\linewidth]{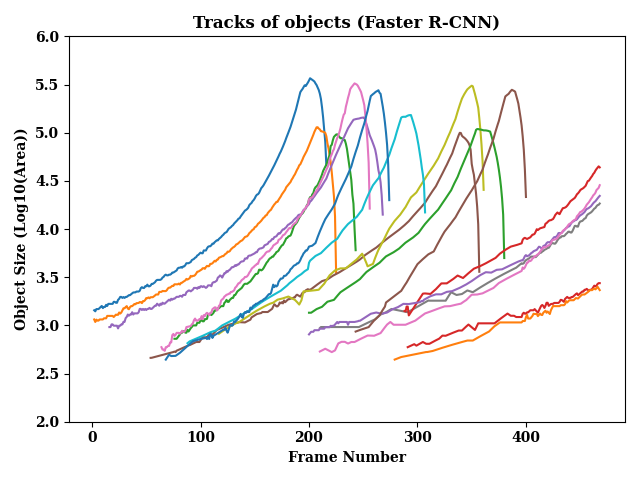}
    \includegraphics[width=0.49\linewidth]{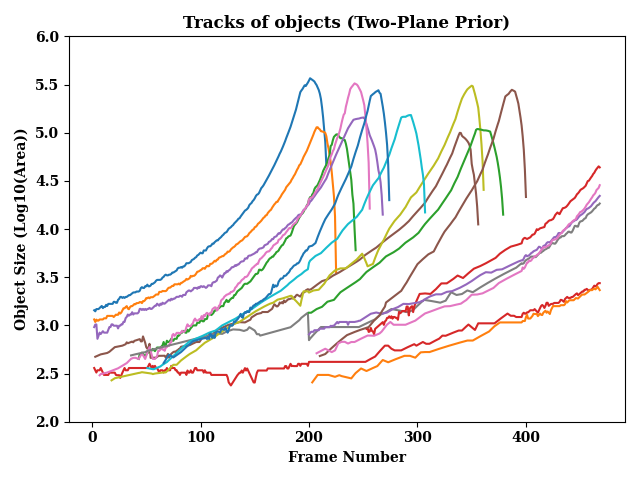}
\caption{\textbf{Tracking Visualization:} To visualize the impact of our two-plane prior, we visualize tracks of length greater than 150 frames tracked by both the methods for a given sequence. We plot object size w.r.t frame numbers (which denotes length). The severe drops of the object sizes for some tracks correspond to nearby object overtaken by our vehicle. We can observe that some objects are detected earlier and are tracked for a longer time.}
    \label{fig:appendix-track-viz-2}
\end{figure*}

\noindent
\textbf{Tracking Visualizations:} We present some tracking visualization in Figures~\ref{fig:appendix-track-viz-1} and~\ref{fig:appendix-track-viz-2}. These visualizations motivate us to define the following metrics.

\noindent
\textbf{Detecting and Tracking for Longer:} We wish to understand if Two-Plane Prior is able to detect an object for a longer lifespan. This is important in autonomous driving situations, wherein we want to detect far-away objects as quickly as possible or any object moving away from us. 

\noindent
Prior tracking quality metrics such as MT\% and ML\% check the ratio of tracks that are mostly tracked or mostly lost. However, this does not capture the track length improvements. We propose to compare the average extension of a track (\textbf{ATE}) compared to the baseline detection method. Given a track $\tau$, $E_{\tau}$ can be positive or negative, and is given by,

\begin{equation}
    E_{\tau}(m, b, gt) = (L_{m} - L_{b}) / L_{gt}
\end{equation}

where $m$ is the method, $b$ is baseline and $gt$ is the ground truth track, while $L$ denotes track length. \textbf{ATE} is the average over tracks across all sequences. However, as the metric weighs all tracks equally, which is unfair for extremely small track lengths, thus, we only consider ground truth tracks which are atleast 5 seconds or 150 frames long.

\noindent
\textbf{Detecting and Tracking Smaller Objects:} Given a track, we wish to observe if Two-Plane Prior is able to detect an object when it's ``smaller'' compared to other methods. This is important in autonomous driving situations, wherein we would like to detect further away objects, which would appear smaller. We wish to compute the minimum object size tracked (\textbf{MOS}). We employ a proxy for object size, $size(x) = log(area(x))$ where $x$ denotes an object bounding box, as the area quadratically increases. For a given ground truth track $\tau$, let $o_\tau$ denote minimum object size of an object, while $O_\tau$ denotes maximum object size. Let $c_\tau$ denote the minimum object size in the predicted track currently considered. We can write,

\begin{equation}
    M_\tau = \frac{c_\tau - o_\tau}{O_\tau - o_\tau}
\end{equation}

$M_\tau$ is averaged over all tracks across all sequences to obtain \textbf{MOS}.

\section{Detection on the Commuter Bus}

\noindent
The bus is equipped with a Jetson AGX edge device. The edge device communicates with a modified onboard-NVR recording bus data from 7 cameras, two inside the bus and five on the outside of the bus. The cameras record data at $5 FPS$ at 720P resolution for $8$ working hours of the bus, totalling 1.08 million frames everyday. It is not feasible to transmit and process this data on the cloud due to bandwidth and compute limitations, and privacy concerns. Thus, the edge device and the NVR are part of a distributed edge-cloud infrastructure wherein the edge device is employed to process these simultaneous streams, only relevant frames are transmitted to cloud machines where we do further offline analysis.

\noindent
We analyze bus streams to build an actionable map of public infrastructure, for instance, which areas need a trash pickup or where does snow needs to be shovelled. We also provide real-time feedback to the bus driver, informing them of people who may need assistance (say, on wheelchairs, or with a stroller or service animal) getting on the bus. Thus we employ an object detector to detect trash cans, garbage bags and people with an assistive device. Our system has to operate at near real-time on all streams simultaneously, rendering cloud-transmission-turn-around infeasible.

\begin{table}[]
{\footnotesize
\centering
\setlength{\tabcolsep}{3pt}
\begin{tabular}{lllllllll}
\toprule
Method                                    & Scale & $AP_{50}$ & $AR$ & $AR_S$ & $AR_M$ & $AR_L$ & Latency (ms) \\
\midrule
Faster R-CNN                               & 0.5x & 45.0 & 31.7 & 0.5 & 37.3 & 46.9 & $154 \pm 8.5$ \\
Two-Plane Prior                  & 0.5x & \textbf{77.2} & \textbf{61.7} & \textbf{16.4} & \textbf{69.9} & \textbf{68.7} & $158 \pm 7.5$\\
\midrule
Faster R-CNN                               & 0.75x & 58.6 & 41.1 & 10.5 & 45.3 & 54.7 & $240 \pm 8.5$ \\
Two-Plane Prior                  & 0.75x & \textbf{84.5} & \textbf{68.3} & \textbf{38.8} & \textbf{72.9} & \textbf{73.3} & $245 \pm 10$ \\
\midrule
\multicolumn{8}{c}{Baseline at higher scales} \\
\midrule
Faster R-CNN  & 1x                         & 68.2 & 41.5 & 16.9 & 47.5 & 57.1 &  $350 \pm 15$ \\
\bottomrule
\end{tabular}
\caption{\textbf{Rare Object Detection on the Commuter Bus:} We compare our approach with a baseline Faster R-CNN. We observe improved precision and recall over the baseline, specially for small and medium sized objects. Do note, for $\approx$1FPS throughput over five simultaneous streams, average latency of $200$ms should be achieved (however, this is not an enforced latency budget for streaming perception~\cite{li2020towards}). } 
\label{table:appendix-bus-data}
}
\end{table}

\noindent
As we employ the edge device to filter out relevant frames, detecting all the objects in the scene is more important than the precision and localization accuracy (a frame once, marked ``relevant'', is sent to cloud where we employ larger models at higher resolutions without constraints).

\noindent
\textbf{Dataset Acquisition:} For research purposes, we do record all the data\footnote{\href{https://what-if.xkcd.com/31/}{Transmission is infeasible, HDD's swapped physically.} (\href{https://en.wikipedia.org/wiki/Sneakernet}{Sneakernet})}, which is humongous ($\approx$30 Terabytes till now) and the instances are rare, we were able to identify 3.5K such frames (temporally subsampled to 750) through a semi-automatic method. Firstly, we only sampled frames from the camera that is facing the sidewalk (people entering the bus are visible). We then geo-fenced images from bus-stop locations and major intersections on the bus route reducing the set to 780K images. Then, we employ off-the-shelf Detic Swin-B Large Faster R-CNN with CLIP (for custom vocabulary)~\cite{zhou2022detecting} and find images with "wheelchair", "stroller", "walker", "crutches", "cane", "dog", "animal", "trolley", "cart", "trash can", "garbage bin", "garbage", "garbage bag" categories with a confidence threshold of 0.25. This model has a high false positive rate for these rare classes, and we were able to automatically filter a set of 21K images, and manually filtered these to yield 3.5K images. As many of these images were part of dense temporal sequences, we further sub-sampled temporally within each sequence yielding 750 samples. We manually annotated these images with object bounding boxes and categories ("trash-can", "garbage-bag" and "person-requiring-assistance"; labels from Detic~\cite{zhou2022detecting} were not accurate). As the data is recorded over the course of a year, we split the train and test test (70\% - 30\%) using the date stamp (images taken on the same day are in the same split) so that the model doesn't overfit. 

\noindent
\textbf{Hardware Platform:} We set the Jetson AGX to consume 30+ Watts (MAXN configuration; no power budget). Memory is measured using the \texttt{tegrastats} utility, while we use Jetpack 4.6.1 and pytorch 1.6, mmdetection 2.7 (+ mmcv 1.15) compiled for Jetson AGX to measure latency consistently across methods (models can be compiled with TensorRT and trained with mixed precision for additional orthogonal improvements).  

\noindent
\textbf{Results:} In this case, just like autonomous driving, we observe that the vanishing point is highly local. Due to overheads of vanishing point estimate on our edge device, we instead employ the average vanishing point, and cache saliency $S$, considerably reducing our approach's latency and memory while maximizing accuracy. From Table~\ref{table:appendix-bus-data}, we observe $AR$ and $mAP_{50}$ for the baseline (Faster R-CNN) and our approach at 0.5x and 0.75x scales. Our method consistently outperforms the baseline method at the same scale, showing both better precision and recall while incurring only \textbf{4ms} additional latency and \textbf{22 MB} memory overheads.

\section{Qualitative Results}

% \blue{Add temporal + geometric}

% \blue{How it works in different Situations: Bus Data on Highways, BDD etc}

\begin{figure*}
    \centering
    \includegraphics[width=1\linewidth]{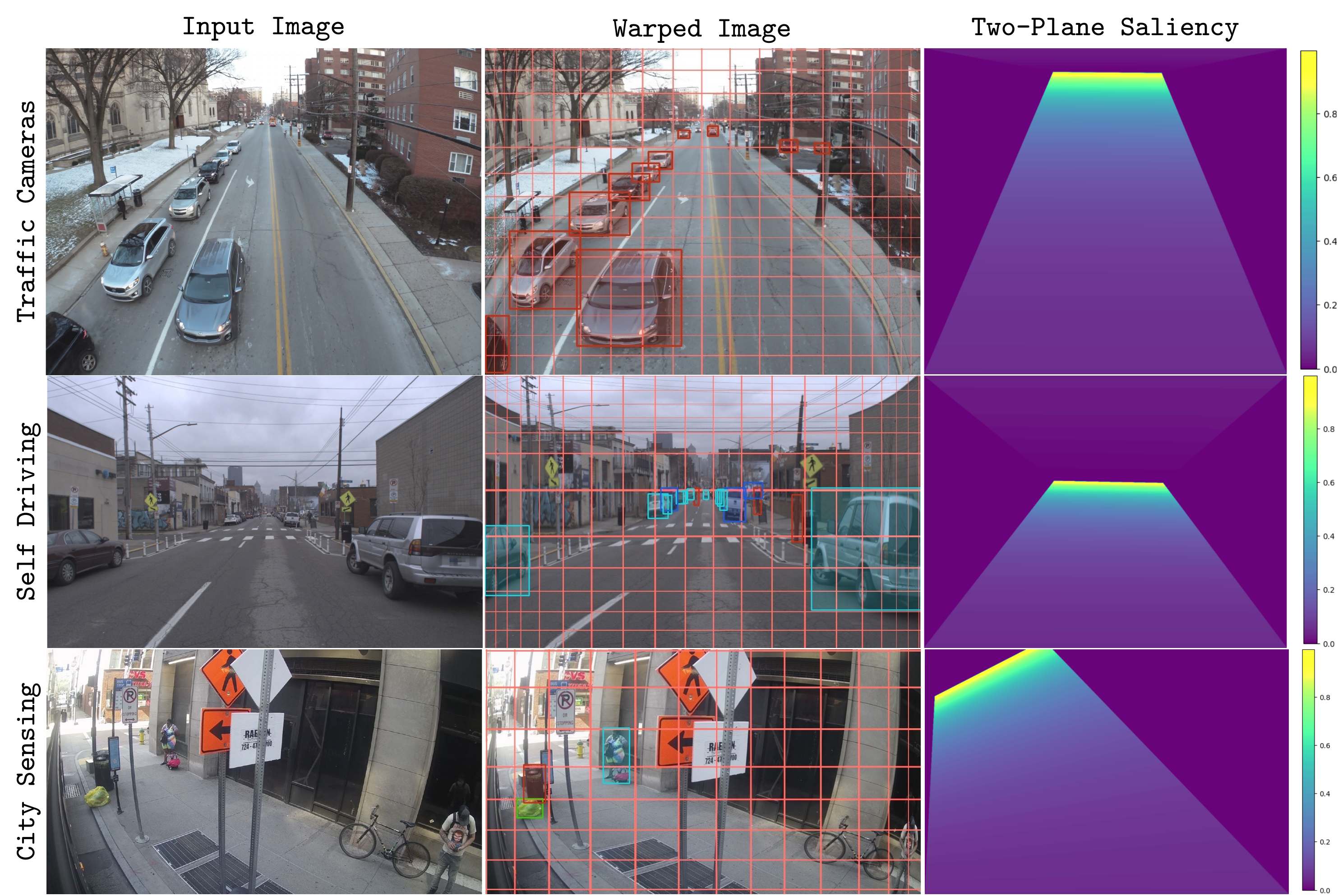}
    \caption{\textbf{Two-Plane Prior Based Warping:} Two-Plane Prior is defined by a few parameters that describe two planar regions in the direction of the vanishing point in the 3D scene (See Section 3.1 and 3.2 in manuscript). Firstly, we can observe the Two-Plane Prior's explicit dependence on the vanishing point $v$ in the saliency maps. Next, as we can observe from grid lines (equidistant in the original image) overlaid on top of the warped images, the extent of spatial warping varies across datasets (WALT, Argoverse-HD and Commuter Bus), showing us the need for learnable parameter $\nu$ over prior work which do not directly model this relationship. Lastly, notice the second plane's effect in sampling. The second plane acts as a "counter-balance" to reduce distortion, and the plane is  \textbf{faintly observable} (contrast adjusted for better visibility).}
    \label{fig:appendix-two-plane}    
\end{figure*}

\begin{figure*}
    \centering
    \includegraphics[width=1\linewidth]{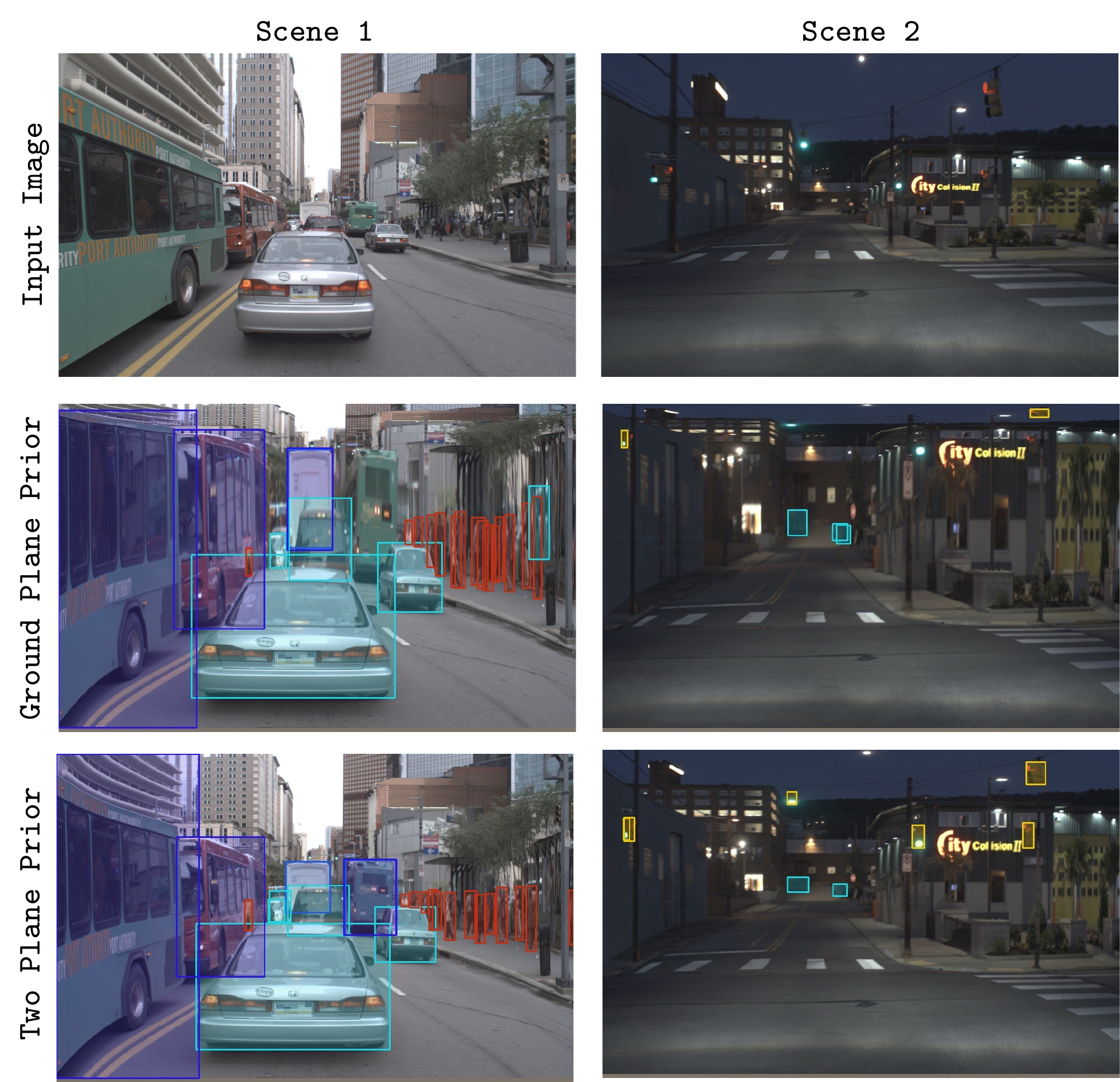}
    \caption{\textbf{Ground Plane Prior vs Two-Plane Prior:} This figure demonstrates how crucial it is to model the second plane. Learning is difficult with Ground Plane Prior (Section 5.1 in the manuscript) and causes heavy distortion of non-ground-plane regions. \\
    \textbf{\textit{Scene 1:}} Detector with Ground Plane Prior misses nearby tall objects because of heavy distortion. Turquoise colored bus on the right (blue box) is detected when Two-Plane prior is used and missed with Ground Plane prior.
    \\ \textbf{\textit{Scene 2:}} Objects not on the ground plane are missed as they are squished by the Ground Plane Prior. Yellow boxes denote the traffic lights. All 6 traffic lights in the scene were detected when Two-Plane prior is used while Ground Plane prior missed 4 traffic lights.}
    \label{fig:appendix-ground-plane}    
\end{figure*}

\begin{figure*}
    \centering
    \includegraphics[width=1\linewidth]{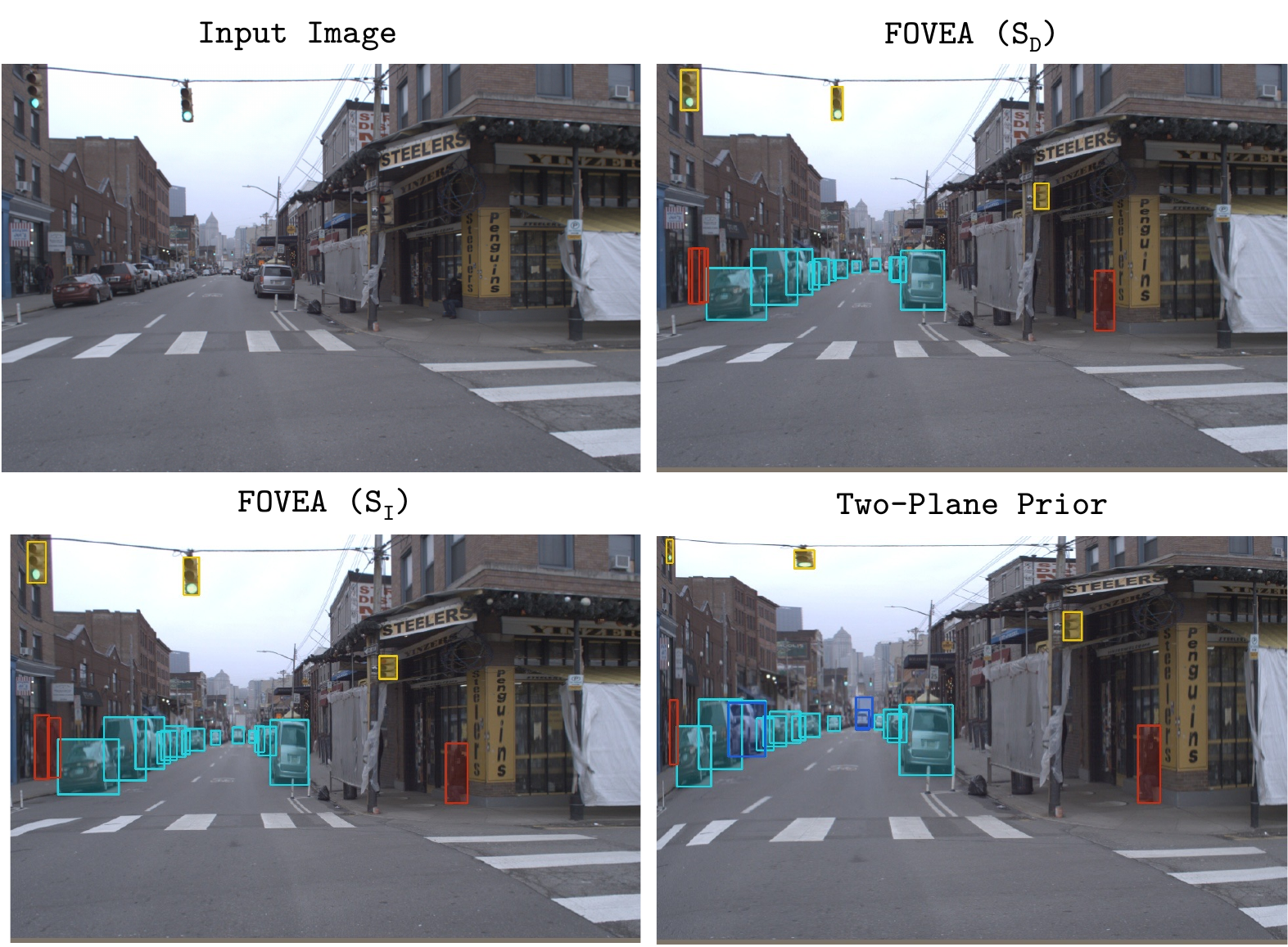}
    \caption{\textbf{Qualitative Comparison with Fovea Warps on Argoverse-HD:} We observe that the reliance on the vanishing point $v$ allows the warp to sample in the direction of the road even while making turns. Far ahead on the road, a $truck$ (dark-blue) is not detected by Fovea ($S_{D}$ or $S_{I}$), but correctly detected by our approach.}
    \label{fig:appendix-warp-compare-1}    
\end{figure*}

\begin{figure*}
    \centering
    \includegraphics[width=1\linewidth]{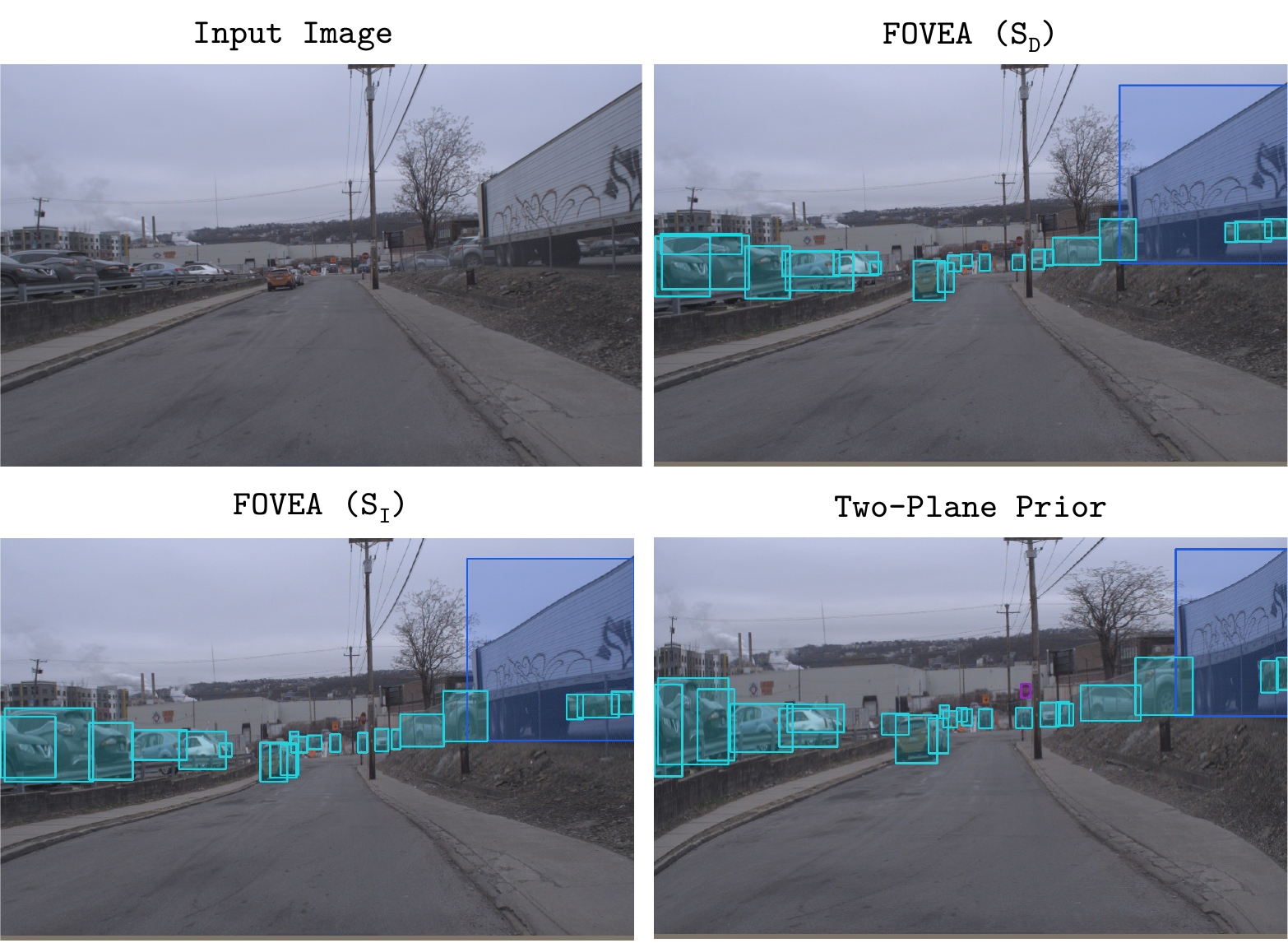}
    \caption{\textbf{Qualitative Comparison with Fovea Warps on Argoverse-HD:} We observe that scale factor $\nu$ models the extent of sampling better. Fovea ($S_D$ or $S_I$) misses the $stop-sign$ (a magenta box in the middle of the image) which our method is able to detect (as it's larger in the warped image). Fovea ($S_{I}$) notes that in their method, regions immediately adjacent to magnified regions are often contracted which is noticed in this case.}
    \label{fig:appendix-warp-compare-2}    
\end{figure*}

\begin{figure*}
    \centering
    \includegraphics[width=1\linewidth]{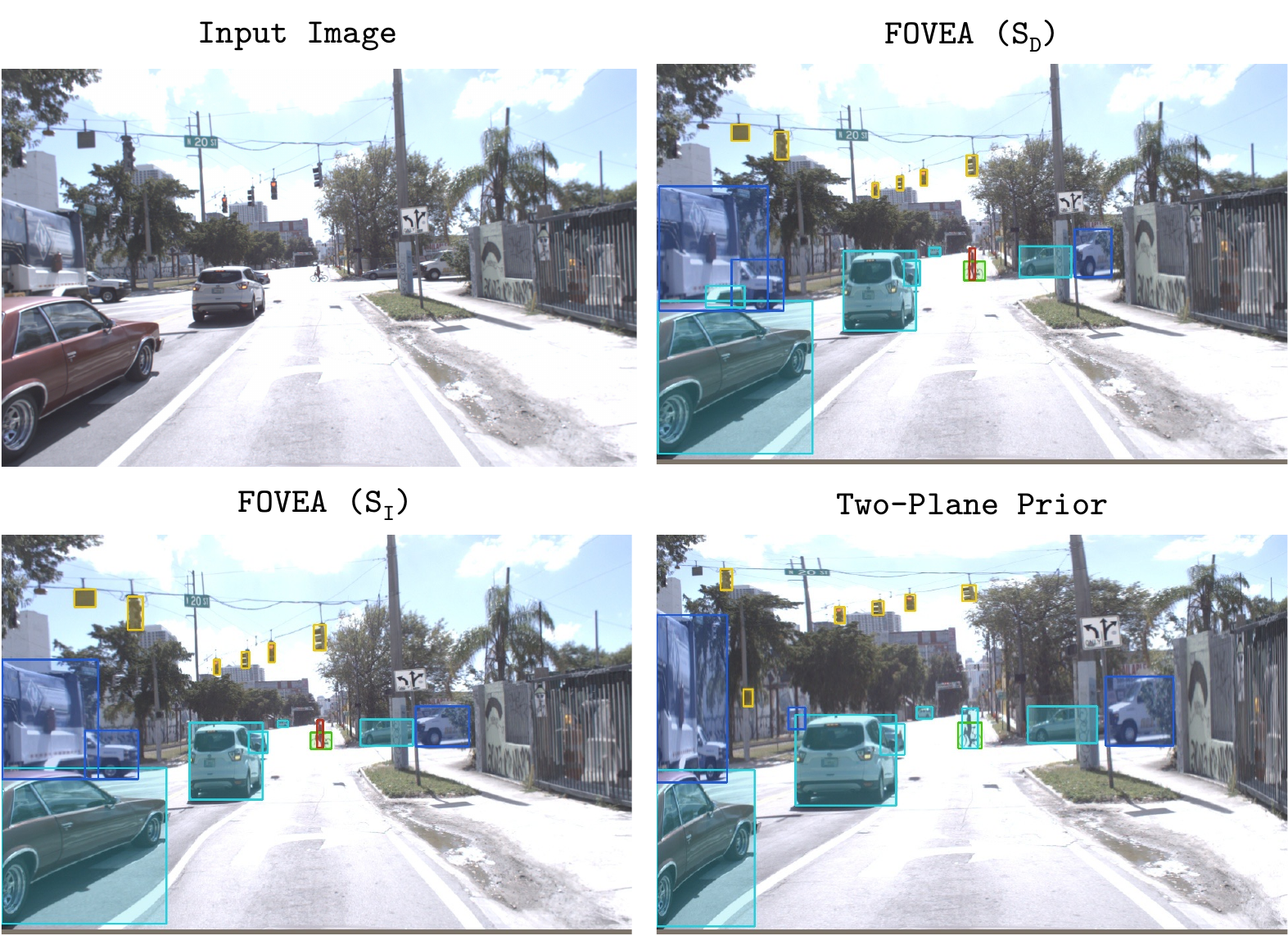}
    \caption{\textbf{Qualitative Comparison with Fovea Warps on Argoverse-HD:} \textit{Failure Case:} The model has misclassified a pedestrian as car in the image warped by the Two-Plane Prior while correctly classified by Fovea ($S_{D}$ or $S_{I}$) (red), likely due to the presence of bicycle and heavier distortion.}
    \label{fig:appendix-warp-compare-3}    
\end{figure*}

\begin{figure*}
    \centering
    \includegraphics[width=1\linewidth]{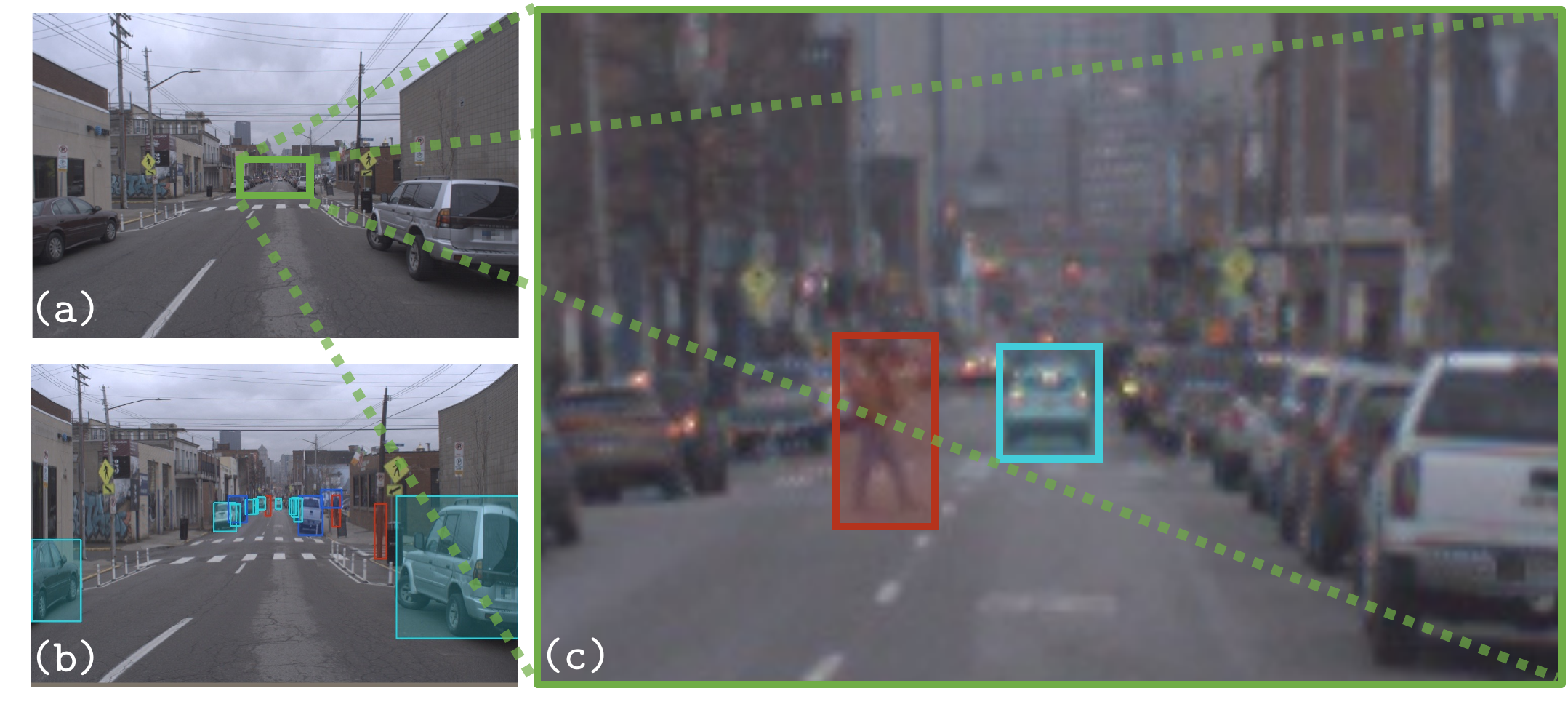}
    \caption{\textbf{Detection of Far Away Objects:} Our Two-Plane Prior boosts the detection of small far-away objects at lower resolutions (depicted image from Argoverse-HD dataset). The cropped green region in the (a) original image is (c) zoomed in while (b) shows all the detections in the warped image. Our method detects far-away pedestrian and car.}
    \label{fig:appendix-far-away-1}
\end{figure*}

\begin{figure*}
    \centering
    \includegraphics[width=1\linewidth]{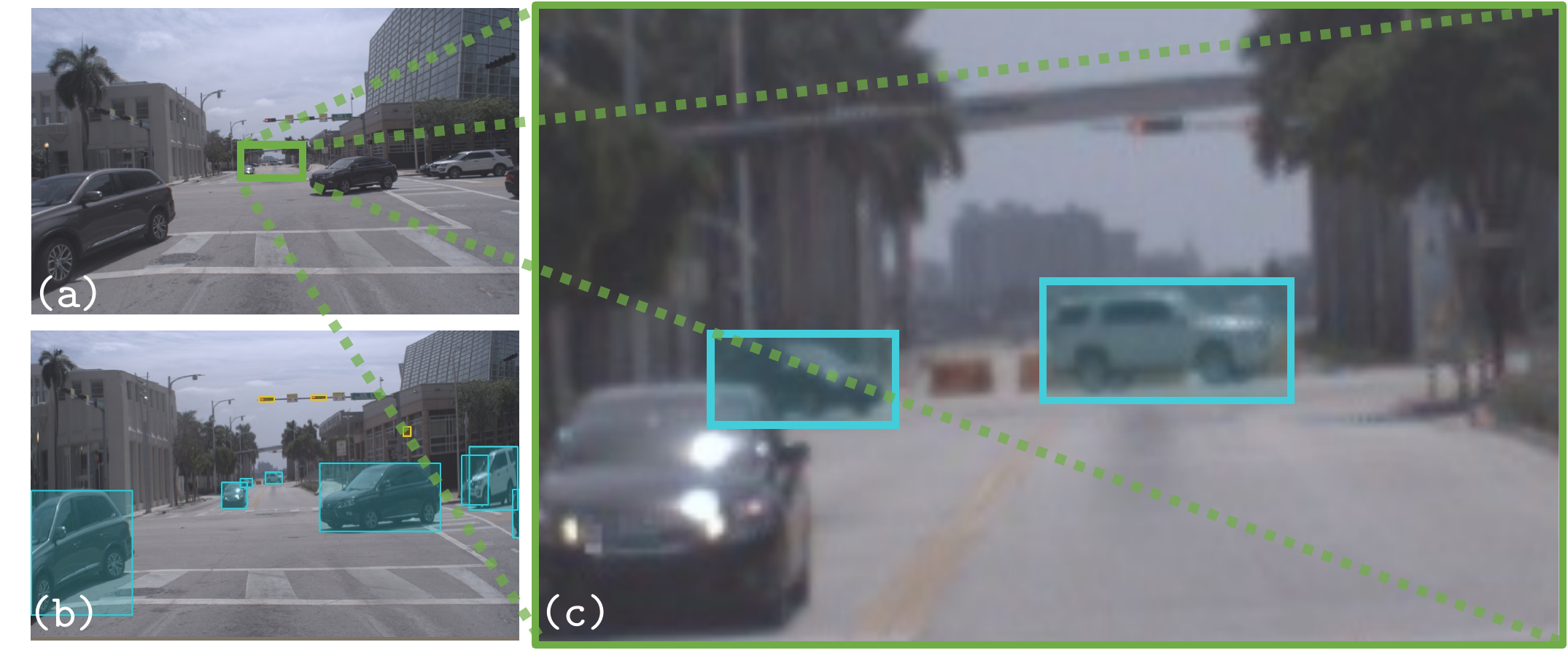}
    \caption{\textbf{Detection of Far Away Objects:} Our Two-Plane Prior boosts the detection of small far-away objects at lower resolutions (depicted image from Argoverse-HD dataset). The cropped green region in the (a) original image is (c) zoomed in while (b) shows all the detections in the warped image. Our method is able to detect the occluded car.}
    \label{fig:appendix-far-away-2}
\end{figure*}

\noindent
We present the variations of our proposed Two-Plane Perspective Prior across different datasets and scenarios in Figure~\ref{fig:appendix-two-plane}. We also show case of the major failure mode of just employing Ground Plane Prior in Fig~\ref{fig:appendix-ground-plane}. We also show a qualitative comparison with prior work in Figures~\ref{fig:appendix-warp-compare-1},~\ref{fig:appendix-warp-compare-2} and ~\ref{fig:appendix-warp-compare-3}. Lastly, we take a closer look at some of the far away objects that were detected in Figures~\ref{fig:appendix-far-away-1} and ~\ref{fig:appendix-far-away-2}. The accompanying website further illustrates some of the aspects of our method.

% {\small
% \bibliographystyle{ieee_fullname}
% \bibliography{egbib}
% }

\end{document}